\documentclass[10pt,twocolumn,letterpaper]{article}

\usepackage{iccv}              

\usepackage{algorithm}  
\usepackage{algpseudocode}  
\usepackage{amsmath}  
\usepackage{graphicx} 
\usepackage{amsthm}  
\theoremstyle{definition}

\newtheorem{lemma}{Lemma}
\usepackage{multirow}
\usepackage{booktabs}
\usepackage{adjustbox}
\usepackage{graphicx}
\usepackage{comment}
\usepackage{amssymb}

\def\xnoi{x_{\textsc{n}}}

\ifodd 1

\newcommand{\com}[1]{\textbf{\color{red} \left(Comment: #1\right) }}
\newcommand{\comg}[1]{\textbf{\color{blue} \left(COMMENT: #1\right)}}
\newcommand{\response}[1]{\textbf{\color{blue} \left(RESPONSE: #1\right)}}
\else

\newcommand{\com}[1]{}
\newcommand{\comg}[1]{}
\newcommand{\response}[1]{}
\fi
\definecolor{iccvblue}{rgb}{0.21,0.49,0.74}
\usepackage[pagebackref,breaklinks,colorlinks,allcolors=iccvblue]{hyperref}

\def\paperID{8225} 
\def\confName{ICCV}
\def\confYear{2025}

\title{MIGA: Mutual Information-Guided Attack on Denoising Models 
 \\ for Semantic Manipulation}

\author{
    Guanghao Li$^{1,2}$\thanks{Equal contribution.} \quad
    Mingzhi Chen$^{2}$\footnotemark[1] \quad 
    Hao Yu$^{1}$\footnotemark[1] \quad
    Shuting Dong$^{1}$\footnotemark[1] \quad
    Wenhao Jiang$^{3}$ \quad \\
    Ming Tang$^{2}$\thanks{Corresponding author: \texttt{tangm3@sustech.edu.cn}} \quad
    Chun Yuan$^{1}$\thanks{Corresponding author: \texttt{yuanc@sz.tsinghua.edu.cn}} \\
    $^{1}$SIGS, Tsinghua University \quad
    $^{2}$Southern University of Science and Technology \\
    $^{3}$Guangdong Laboratory of Artificial Intelligence and Digital Economy (SZ) \\
}

\begin{document}
\maketitle

\begin{abstract}

Deep learning-based denoising models have been widely employed in vision tasks, functioning as filters to eliminate noise while retaining crucial semantic information. Additionally, they play a vital role in defending against adversarial perturbations that threaten downstream tasks. However, these models can be intrinsically susceptible to adversarial attacks due to their dependence on specific noise assumptions. Existing attacks on denoising models mainly aim at deteriorating visual clarity while neglecting semantic manipulation, rendering them either easily detectable or limited in effectiveness.
In this paper, we propose Mutual Information-Guided Attack (MIGA), the first method designed to directly attack deep denoising models by strategically disrupting their ability to preserve semantic content via adversarial perturbations. By minimizing the mutual information between the original and denoised images—a measure of semantic similarity—MIGA forces the denoiser to produce perceptually clean yet semantically altered outputs. While these images appear visually plausible, they encode systematically distorted semantics, revealing a fundamental vulnerability in denoising models. These distortions persist  in denoised outputs and can be quantitatively assessed through downstream task performance. We propose new evaluation metrics and systematically assess MIGA on four denoising models across five datasets, demonstrating its consistent effectiveness in disrupting semantic fidelity. 
 Our findings suggest that denoising models are not always robust and can introduce security risks in real-world applications.

\end{abstract}

\begin{figure}[ht]
    \centering
    \includegraphics[width=0.48\textwidth]{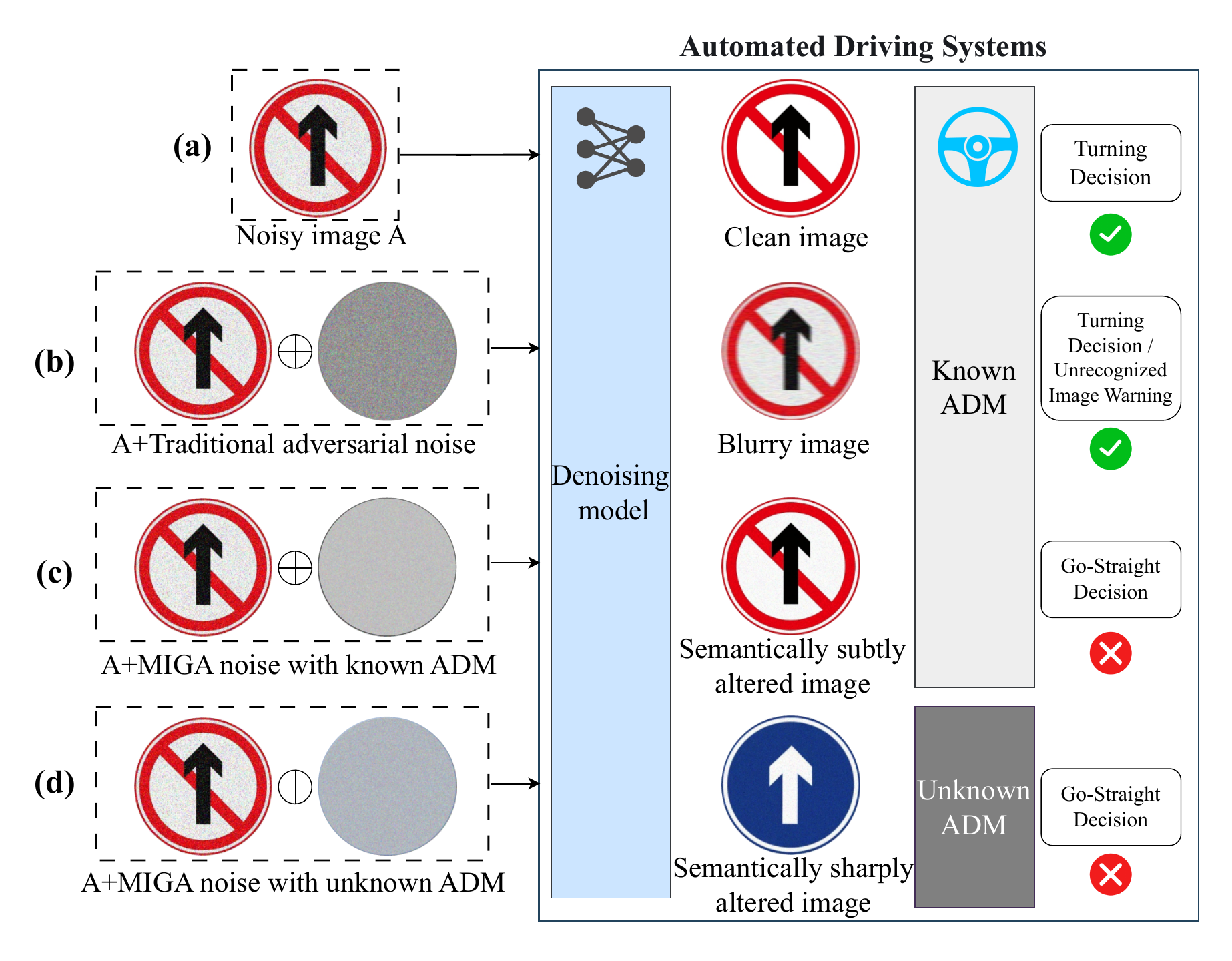} 
    \vspace{-1cm}

\caption{Adversarial attacks on denoising models in automated driving systems.  
A noisy ``No Straight'' sign \(A\) may contain both natural noise and adversarial perturbations designed to mislead Automated Driving Model (ADM). (a) The denoiser typically restores it to a clean and correct form, allowing the ADM to make the correct ``Turning Decision''.  
(b) Traditional attacks on denoising models degrade image quality, making it blurry but rarely altering semantics. This often results in either a correct decision or an ``Unrecognized Image Warning''.  
(c) Our MIGA attack, with knowledge of the ADM, subtly alters semantics to induce an incorrect ``Go-Straight Decision''.  
(d) Without ADM knowledge, MIGA forces the denoised image’s semantics to resemble a reference ``Go Straight'' sign, leading to a wrong decision.}

    \vspace{-0.7cm}

    \label{fig:overview}
\end{figure}

\section{Introduction}
\label{sec:intro}

Image denoising is a fundamental task in computer vision, aiming to recover visually clean images from noisy observations~\cite{rudin1992nonlinear, buades2005non}. Beyond restoring image clarity, modern denoising models also strive to preserve \emph{critical semantic information} vital for downstream tasks such as autonomous driving~\cite{guoqiang2023bilateral, kloukiniotis2022countering}, medical diagnosis~\cite{kaur2018review, jifara2019medical}, and remote sensing~\cite{rasti2021image,singh2021novel}. Moreover, denoising has been adopted as a defense mechanism to remove adversarial noise before it reaches the final model~\cite{liao2018defense, bakhti2019ddsa}. Recently, the advent of deep learning has led to significant advancements in denoising models, setting new benchmarks across various datasets.~\cite{zhang2017beyond, zhang2021plug, zhang2019residual, li2023efficient, liang2021swinir, zamir2022restormer}.

However, despite these advances, deep learning-based denoising models remain inherently vulnerable and can become targets of adversarial attacks, compromising two key aspects of their outputs: clarity and semantic fidelity ~\cite{ning2023evaluating}.
Specifically, clarity refers to low-level perceptual quality, which is typically assessed using image quality metrics such as PSNR and SSIM~\cite{wang2002universal, goodfellow2014explaining}, while semantic fidelity refers to the preservation of high-level content, which needs to be evaluated based on downstream task performance.  Traditional adversarial attacks primarily aim to degrade clarity metrics , but they often ignore manipulating semantics in a targeted manner~\cite{szegedy2013intriguing}. Additionally, these attacks frequently introduce visible artifacts, making them easily detectable. These limitations reduce their effectiveness in real-world scenarios. As illustrated in ~\cref{fig:overview}, in automated driving systems that integrate denoising models, traditional adversarial attacks primarily degrade image quality while leaving semantic content largely unaffected. As a result, the Automated Driving Model (ADM) can still recognize the intended meaning, leading to the correct ``Turning Decision''. Moreover, if the image degradation is too severe, the system may trigger an ``Unrecognized Image Warning'', preventing an incorrect decision altogether.  

\begin{figure*}[ht]
\vspace{-0.6cm}
    \centering
    \includegraphics[width=0.9\textwidth]{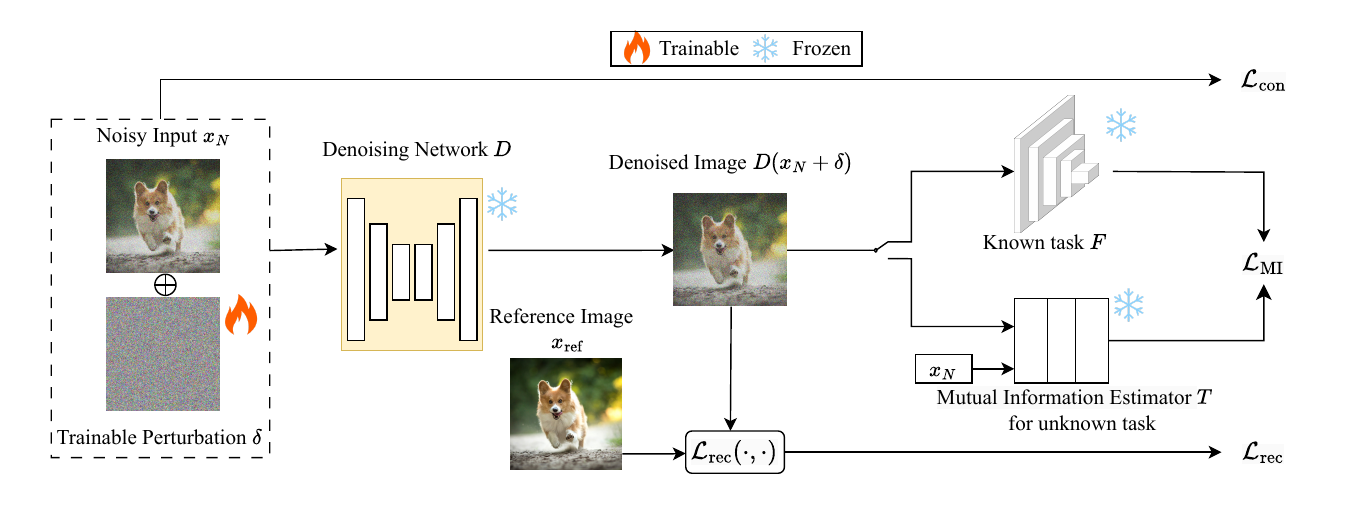} 
    \vspace{-0.6cm}
    \caption{Architecture overview of MIGA. $\mathcal{L}_{\text{con}}$ enforces the imperceptible perturbation constraint; $\mathcal{L}_{\text{rec}}$ ensures image clarity using a clean reference image $x_{\text{ref}}$; $\mathcal{L}_{\text{MI}}$ characterizes the task-relevant mutual information. When the downstream task is known, the mutual information is estimated using $F$; when the downstream task is unknown, it is estimated using a mutual information estimator $T$.}
    \vspace{-0.4cm}

    \label{fig:overview_pipeline}
    
\end{figure*}

To enhance the effectiveness of adversarial attacks on denoising models in real-world scenarios, we identify two key challenges. \textbf{(1)} The attack should avoid introducing noticeable artifacts on denoised images, preserving a high-quality visual appearance that is less likely to be flagged or rejected. \textbf{(2)} More importantly, it must subtly alter the semantic content rather than merely degrading image clarity, leading to semantic discrepancies that affect downstream model predictions.
To address these challenges, we propose the \emph{Mutual Information-Guided Attack} (MIGA),  a novel adversarial framework that introduces imperceptible perturbations to manipulate semantic features during the denoising process by minimizing the mutual information (MI) \cite{shannon1948mathematical} between the original and denoised images.  MI quantifies the amount of shared semantic content between two variables, and in the context of denoising, it specifically reflects the extent to which semantic information is retained after noise removal \cite{zhu2005new}.  
 By introducing carefully crafted perturbations, MIGA minimizes MI through a mutual information loss  jointly optimized with a perturbation constraint loss and a reconstruction loss, as illustrated in \cref{fig:overview_pipeline}.
 This induces the denoiser to generate visually clean images with systematically altered semantics, ultimately undermining its reliability and propagating corrupted semantics to downstream tasks.

We further distinguish between known and unknown downstream tasks. A task is known if the attacker has prior knowledge of its objective and model architecture, enabling tailored perturbations. Otherwise, it is unknown, making targeted semantic shifts challenging.
To address this, MIGA adapts its strategy accordingly.  In the known setting, it analyzes the impact of semantic shifts on task performance to craft perturbations that make the denoiser produce a visually clean yet subtly altered image. In the unknown setting, the lack of task-specific guidance prevents direct optimization. To overcome this, MIGA introduces a reference image to guide semantic alignment, enabling controlled semantic modifications without explicit knowledge of the downstream model.
As shown in Fig.~\ref{fig:overview}, in the known scenario, MIGA manipulates the denoiser to generate a denoised ``No Straight'' sign that appears visually clean but subtly alters its semantics, leading to an incorrect ``Go Straight Decision''. Meanwhile, in the unknown scenario, MIGA uses a “Go Straight'' reference image to induce semantic shifts, causing wrong downstream decisions.

In summary, our main contributions include:

\begin{itemize}[leftmargin=*]
    \item \textbf{First semantic adversarial attack on denoising models.} We propose MIGA to specifically disrupt \emph{semantic fidelity} while preserving high visual quality, revealing a previously overlooked vulnerability in deep denoising models.
    \item \textbf{Task-relevant MI formulation.} We introduce  MI-based loss terms to systematically reduce the task-relevant semantic overlap between the original and denoised images, effective in both known and unknown task scenarios.
    \item \textbf{Extensive experiments \& tailored metrics.} We validate MIGA on four denoising models and five datasets, designing new evaluation metrics to measure semantic manipulation. Results confirm MIGA’s ability to produce perceptually clean yet semantically disruptive outputs that mislead real-world downstream tasks.
\end{itemize}


\section{Related work}

\textbf{Image denoising.}  Image denoising is essential for applications such as autonomous driving~\cite{guoqiang2023bilateral, kloukiniotis2022countering}, medical diagnosis~\cite{kaur2018review, jifara2019medical}, and remote sensing~\cite{rasti2021image,singh2021novel}, where clear visual data is crucial. Moreover, denoising has been widely adopted as a defense mechanism to remove adversarial noise before it reaches the final model~\cite{liao2018defense, bakhti2019ddsa}. Traditional denoising approaches such as total variation \cite{rudin1992nonlinear}, Non-Local Means \cite{buades2005non}, and sparse coding \cite{mairal2009non} established early foundations. With the advent of deep learning, performance has significantly improved using CNN- \cite{zhang2017beyond}, non-local \cite{zhang2019residual}, and Transformer-based \cite{liang2021swinir, zamir2022restormer} architectures. Recent studies emphasize that preserving semantic information during denoising is critical for downstream task performance \cite{buchholz2020denoiseg}. However, despite these advances, deep learning-based denoising models remain susceptible to adversarial attacks, which can compromise their ability to maintain both image clarity and semantic fidelity~\cite{ning2023evaluating}.

\textbf{Mutual information.} 
MI quantifies the shared information between two variables \cite{shannon1948mathematical} and is widely utilized in tasks such as image registration \cite{maes1997multimodality}, feature selection \cite{peng2005feature}, and representation learning \cite{hjelm2018learning}. In the context of image denoising, MI serves as a measure of dependency between noisy and clean images \cite{zhu2005new}, guiding models to retain essential semantic information while removing noise. For example, denoising autoencoders leverage MI to learn robust representations that preserve semantic content \cite{vincent2010stacked}. However, prior work has primarily focused on maximizing MI for semantic preservation, while its potential for adversarial manipulation in denoising remains largely unexplored.

\textbf{Adversarial attacks on denoising models.} Adversarial attacks introduce subtle, often imperceptible perturbations into input images, forcing models to produce incorrect outputs \cite{goodfellow2014explaining}. While adversarial robustness has been extensively studied in classification models, recent research highlights that denoising models are also susceptible to adversarial attacks \cite{ryou2024robust, ning2023evaluating}. Traditional attacks on denoising models primarily disrupt entropy reduction, aiming to degrade image clarity \cite{agnihotri2023cospgd, ijcai2022p211}, but they often generate visible artifacts, making them more detectable. Furthermore,  these attacks generally neglect the manipulation of semantic information during denoising. A notable exception is Pasadena \cite{cheng2021pasadena}, which designs specific denoising models to craft misleading adversarial examples. However, its relies on customized architectures, limiting its generalizability to standard denoising models. In contrast, we propose a task-agnostic framework that strategically manipulates mutual information to generate imperceptible yet semantically altered adversarial examples. Our approach does not require architectural modifications, making it broadly applicable across diverse denoising models and downstream tasks.

\section{Problem setup}\label{sec:problem}

Consider a clean image $x\in \mathbb{R}^n$. This image is corrupted by noise $\eta$ and yields a noisy image $ \xnoi = x + \eta$. The clean image $x$ is unknown, while the noisy image $\xnoi$ is given. There is a denoising model $D$ that aims to recover the clean image from the noisy image $\xnoi$ such that $D(\xnoi) \approx x $. After the image has been denoised, there is a downstream task (e.g., classification, regression, human perception) that uses this denoised image to determine certain outputs (e.g., label, regression outcome). Let $F$ denote the downstream task. Let $C$ denote the ground-truth output of downstream task $F$ by inputting the clean image $x$, i.e., $F(x) = C$. 

In this work, we aim to find an adversarial perturbation $\delta$ added to the noisy image $\xnoi$, resulting in a modified image $\xnoi + \delta$, such that the denoised output $D(\xnoi + \delta)$ retains high clarity but alters semantic content to mislead the downstream task model $F$. Formally, our goal is to find an adversarial perturbation $\delta$ that satisfies the following objectives:
\begin{enumerate}
    \item \textbf{Imperceptible Perturbation}: $\delta$ should be small and imperceptible to human observers, ensuring that $\xnoi + \delta$ is visually indistinguishable from $\xnoi$.
    \item \textbf{Image Clarity Preservation}: The denoised image $D(\xnoi + \delta)$ should maintain high visual quality, appearing as a clean image without noticeable artifacts.
    \item \textbf{Semantic Alteration}: The semantic information in $D(\xnoi + \delta)$ should be altered to mislead the downstream task model $F$, i.e., $F(D(\xnoi + \delta))$ differs from $ C$.
\end{enumerate}

In this work, we consider both cases where the downstream task model $F$ is known and unknown.

\section{Methodology}
In this section, we first propose a novel task-relevant mutual information that characterizes the degree of matching between the clean image $x$ and the denoised image $D(\xnoi+ \delta)$, which will be used for quantifying the semantic alteration. Unlike the traditional mutual information metric, our proposed metric can evaluate how much downstream task-relevant information is preserved in image denoising.  
Then, we propose the Mutual Information Guided  Attack (MIGA) to attack denoising models by designing three carefully-crafted loss functions and the optimization process.

\subsection{Task-Relevant mutual information}\label{subsec:mutual}
Traditional mutual information $ I(x; D(\xnoi+ \delta))$ measures the shared information between the original image $x$ and the denoised output {$D(\xnoi + \delta)$}, considering the adversarial perturbation. This metric, however, encompasses both downstream task-relevant and irrelevant details. In this work, to emphasize on the semantic features that are important for the downstream task, we introduce the concept of task-relevant mutual information. This metric quantifies the shared information between the original image $x$ and the denoised image $D(\xnoi+ \delta)$, conditioned on the ground-truth output $ C$ of the downstream task. For example, in an image of a landscape containing a person, the task-relevant mutual information for location recognition would primarily focus on the background features, whereas for face recognition, it would emphasize facial details. Specifically, task-relevant mutual information is defined as follows:
\begin{equation}\label{eq:H}
I(x; D(\xnoi+\delta) \mid C) 
= H(C \mid x) - H(C \mid D(\xnoi+\delta)),
\end{equation}
where \( H(C \mid x) \) and \( H(C \mid D(\xnoi+\delta)) \) represent the conditional entropies of the ground-truth \( C \) given the original and denoised images, respectively. 

This formulation helps us assess how much task-relevant information is preserved in the denoised image. 
Thus, we can achieve semantic alteration by minimizing task-relevant mutual information $I(x; D(\xnoi+\delta) \mid C)$. However, directly computing $I(x; D(\xnoi+\delta) \mid C)$ is impossible because the clean image $x$ is unknown and $H(C \mid D(\xnoi+\delta))$ involves high-dimensional probability distribution estimation. In the following, we present methods to measure or approximate the  task-relevant mutual information for known and unknown downstream task scenarios, where these measures will be used in loss function design.

\textbf{Known downstream task.} 
When the downstream task model $F$ is known, we propose to use the cross-entropy loss $\mathcal{L}_{\text{CE}}(F(D(\xnoi + \delta)), C)$ to characterize the task-relevant mutual information. We prove that maximizing such a cross-entropy loss is equivalent to minimizing the task-relevant mutual information $I(x; D(\xnoi+\delta) \mid C)$, 
with a detailed proof  provided in ~\cref{sec:proof}.

\textbf{Unknown downstream task.}  
When the downstream task \( F \) is unknown, we propose a semantic alteration strategy to estimate the task-relevant mutual information without requiring explicit task-specific knowledge. Specifically, we first construct a dataset of paired images \( (x, x_{\text{alt}}) \), where \( x_{\text{alt}} \) serves as a reference image with altered semantics compared to the original image \( x \).
 The alteration process involves modifying key semantic regions of \( x \), such as objects, scenes, or textual content, which alters the image’s semantics. This paired dataset is then used to pre-train the MINE network \cite{belghazi2018mine}. The network learns to estimate the task-relevant mutual information by modeling the joint distribution between the original image \( x \) and its altered version \( x_{\text{alt}} \), enabling it to capture relevant semantic information without the need for explicit task labels.

Specifically, we train a neural network estimator $T$ to distinguish between positive pairs $(x, D(\xnoi))$ and negative pairs $(x, x_{\text{alt}})$. We use a contrastive loss to encourage  the model to assign higher similarity scores to positive pairs and lower scores to negative pairs, focusing on the model on the image's semantically relevant content. After training, \( T(x, D(\xnoi)) \) measures the similarity between the original image \( x \) and its denoised version \( D(\xnoi) \), preserving the core structure and semantics of the image. By optimizing the adversarial perturbation $\delta$ to minimize $T(x, D(\xnoi+\delta)) $,  task-relevant mutual information can be approximately minimized.

\subsection{Loss functions}

According to the objectives in  \cref{sec:problem}, we propose to consider three losses: $\mathcal{L}_{\text{con}}(\xnoi, \xnoi + \delta)$ enforces the imperceptible perturbation constraint; $\mathcal{L}_{\text{rec}}(x_{\text{ref}}, D(\xnoi + \delta))$ ensures image clarity using a clean reference image $x_{\text{ref}}$;  $\mathcal{L}_{\text{MI}}$ characterizes the task-relevant mutual information. In other words, the objective can be transformed into optimizing the adversarial perturbation $\delta$ such that 
\begin{equation}
\begin{aligned}
 \delta^* = &\arg \min_{\delta} \Big[ \lambda_{\text{con}} \mathcal{L}_{\text{con}}(\xnoi, \xnoi + \delta) \\
& + \lambda_{\text{rec}} \mathcal{L}_{\text{rec}}(x_{\text{ref}}, D(\xnoi + \delta)) + \lambda_{\text{MI}} \mathcal{L}_{\text{MI}} \Big].
\end{aligned}
\end{equation}
We omit the explicit constraint  $\|\delta\|_\infty \leq \epsilon$, because the perturbation  loss $\mathcal{L}_{\text{con}}(\xnoi, \xnoi + \delta)$ has already enforced the imperceptibility of $\delta$.

To implement this objective, we design three carefully-crafted loss functions to guide the optimization process. 

\textbf{Perturbation constraint loss.}
To ensure imperceptibility perturbation, we define the following loss function:
\begin{multline}\label{eq:con}
\mathcal{L}_{\text{con}}(\xnoi, \xnoi + \delta) = \ \alpha \left\| \xnoi - (\xnoi + \delta) \right\|_2^2 \\
+ (1 - \alpha) \mathcal{L}_{\text{perc}}(\xnoi, \xnoi + \delta),
\end{multline}
where \( \mathcal{L}_{\text{perc}}(\xnoi, \xnoi + \delta) \) is the perceptual loss, computed using feature maps from a pre-trained VGG16 network \cite{simonyan2014very}, and \( \alpha \in [0,1] \) controls the trade-off between pixel-level and perceptual similarities.

\textbf{Reconstruction loss.}
The reconstruction loss ensures that $D(\xnoi + \delta)$ remains visually similar to a reference image $x_{\text{ref}}$. Its expression is given as follows:
\begin{multline}
\label{eq:reconstruction_loss}
\mathcal{L}_{\text{rec}}(x_{\text{ref}}, D(\xnoi + \delta)) = \  \beta \left\| x_{\text{ref}} - D(x_n + \delta) \right\|_2^2 \\
 + (1 - \beta) \mathcal{L}_{\text{perc}}(x_{\text{ref}}, D(\xnoi + \delta)),
\end{multline}
where $\beta \in [0,1]$ balances the pixel-wise and perceptual similarities.
The reference image \( x_{\text{ref}} \) is selected based on the downstream task: for a known downstream task, \( x_{\text{ref}} = D(\xnoi) \approx x \); for an unknown downstream task, {\( x_{\text{ref}} = x_{\text{alt}} \)}.

\textbf{Mutual information loss.} The mutual information loss \( \mathcal{L}_{\text{MI}} \) is designed to quantify the task-relevant mutual information, as described in \cref{subsec:mutual}. 

For a \textit{Known Downstream Task}, the mutual information loss is formulated as:
\begin{equation}\label{eq:mi-knonw}
\mathcal{L}_{\text{MI}} = - \mathcal{L}_{\text{CE}}(F(D(\xnoi + \delta)), C),
\end{equation}
where \( \mathcal{L}_{\text{CE}} \) is the cross-entropy loss. 

For an \textit{Unknown Downstream Task}, the mutual information loss is given by
\begin{equation}\label{eq:mi-unknown}
\mathcal{L}_{\text{MI}} =  T(x, D(\xnoi+\delta)),
\end{equation}
where \( T(x, D(\xnoi+\delta)) \) outputs a similarity score between \( x \) and \( D(\xnoi+\delta) \). Minimizing this score reduces the task-relevant mutual information.

\begin{algorithm}[t]
\caption{Training Process of MIGA}
\label{alg:MIGA} 
\begin{algorithmic}[1]
\State \textbf{Input:} Noisy image \( \xnoi = x + \eta \), clean image \( x \), denoising model \( D\), task-specific reference \( x_{\text{ref}} \), perturbation initialization \( \delta = 0 \)
\State \textbf{Output:} Adversarial perturbation \( \delta \)
\While{not converged}
    \State Create adversarial image {\( x_{\text{MIGA}} = \xnoi + \delta \)};
    \State Obtain denoised output {\( y = D(x_{\text{MIGA}}) \)};
    \State Compute  \( \mathcal{L}_{\text{con}}( \xnoi, x_{\text{adv}}) \)  using \eqref{eq:con};
    \State Compute $ \mathcal{L}_{\text{rec}}(x_{\text{ref}}, y) $ using \eqref{eq:reconstruction_loss};
    \State Compute $\mathcal{L}_{\text{MI}}$ using \eqref{eq:mi-knonw} for known task and using \eqref{eq:mi-unknown} for unknown task;
 
    \State \( \mathcal{L}_{\text{total}} = \mathcal{L}_{\text{con}}( \xnoi, x_{\text{adv}})  + \mathcal{L}_{\text{rec}}(x_{\text{ref}}, y) + \mathcal{L}_{\text{MI}} \);
    \State Update perturbation \( \delta \leftarrow \delta - \alpha \nabla_{\delta} \mathcal{L}_{\text{total}} \);
\EndWhile
\end{algorithmic}

\end{algorithm}

\subsection{Training procedure}

We illustrate our overall framework in  \cref{fig:overview_pipeline}, with the detailed process shown in \cref{alg:MIGA}. To optimize the perturbation \( \delta \), we employ a gradient-based approach. We start with a noisy image  \( \xnoi = x + \eta \), where \( x \) is the clean image and \( \eta \) is the added noise. The perturbation \( \delta \) starts at zero and is iteratively updated until convergence.

In each iteration, the adversarial image \( x_{\text{MIGA}} = \xnoi + \delta \) is created. The denoised output \( y = D(x_{\text{MIGA}}) \) is obtained using the given pre-trained denoising model \( D \). The total loss is a weighted sum of the perturbation constraint loss, reconstruction loss, and  mutual information loss, computed based on the task. For known tasks, cross-entropy loss is used. For unknown tasks, mutual information is estimated. The perturbation \( \delta \) is updated through gradient descent with respect to the total loss, with $\alpha$ being the step size.

\section{Experiments}
In this section, we evaluate MIGA on denoising models through experiments with both known and unknown downstream tasks. We describe the datasets, metrics, models, and baselines, followed by an ablation study to assess the contribution of different components. Finally, we analyze the robustness of our adversarial examples against various defense strategies. Further details of our experiments, including experimental settings and implementation, are provided in ~\cref{sec:details}, while ~\cref{more results} presents additional analyses, such as hyperparameter selection and transferability evaluation.

\subsection{Experiments setting}
\textbf{Datasets.}
To evaluate our method, we define specific dataset requirements based on the nature of downstream tasks, which performence serve as indicators for semantic modification. For \textbf{known downstream tasks}, a well-defined task specification and the availability of high-performing pre-trained models are essential. We use the ImageNet-10 dataset \cite{russakovsky2015imagenet}, which consists of a 10-class classification task and is supported by several state-of-the-art pre-trained models.  For \textbf{unknown downstream tasks}, a clean reference image with modified semantics is required. 
We employ the Tampered-IC13 dataset \cite{wang2022detecting}, which includes real-world text alterations in natural scenes, and the MAGICBRUSH dataset \cite{zhang2024magicbrush}, originally designed for image editing tasks, containing both the original images, editing instructions, and the modified results.
To further enhance the generalizability of our evaluation, we incorporate two synthetic datasets. The first is a stylized dataset created using the neural style transfer method \cite{gatys2015neural}, which includes both the original images and their corresponding stylized reference images. Additionally, we create a synthetic text alteration dataset by selecting a subset of images from the SSDI dataset \cite{abdelhamed2018high} as backgrounds, and generating pairs of small but semantically distinct text fragments, placed at the same location in each background.
In the experiment, Gaussian noise was added to the images in different datasets to generate noisy versions. Specifically, a Gaussian noise with a variance of 50 was applied to the ImageNet-10 dataset, while Gaussian noise with a variance of 25 was added to other datasets. In the calculation of \( \mathcal{L}_{\text{rec}} \) during the experiment, the reference image \( x_{\text{ref}} \) used was the clean image, without any added noise.

\textbf{Evaluation metrics}.
To comprehensively evaluate our algorithm, we define a set of metrics targeting three key objectives: (1) the imperceptibility of perturbations, (2) the clarity of denoised images, and (3) the semantic integrity of denoised images.
For \textbf{imperceptibility}, we compute the LPIPS score \cite{zhang2018unreasonable} between the noisy image \(\xnoi\) and the perturbed image \(\xnoi + \delta\). LPIPS measures the perceptual similarity between images, with lower values indicating higher visual similarity. We use $\text{LPIPS}_{\text{con}}$  to denote this perceptual similarity constraint.
For \textbf{clarity}, we measure the similarity between the denoised image and the clean reference using multiple indicators: PSNR, SSIM \cite{wang2002universal}, LPIPS \cite{zhang2018unreasonable}, and image entropy \cite{prashanth2023image}. Higher PSNR and SSIM values, combined with lower LPIPS and entropy scores, reflect better denoising quality and improved image clarity.
To evaluate \textbf{semantic modification}, we employ different approaches based on whether the downstream task is known. 

For \textbf{known downstream tasks}, performance degradation signals semantic changes. For example, in the  classification task, a drop in accuracy suggests potential semantic alterations. For \textbf{unknown downstream tasks}, we assess semantic fidelity by measuring core performance changes in the reference image. In the \textbf{text tampering} dataset, OCR \cite{singh2012survey} is used to extract text from denoised images, and ROUGE-L \cite{lin2004rouge} scores between the extracted and reference text quantify semantic shifts. In \textbf{image editing} datasets, CLIP similarity \cite{radford2021learning} between the generated image and editing instructions gauges adherence to the intended modification. For \textbf{style alteration} tasks, we use Gram matrix loss \cite{gatys2015neural} to compare the denoised image with the stylized reference image, where a lower loss indicates better style alignment.

\textbf{Models and baselines.}
To extensively evaluate our experiments, we select several different denoising network architectures, including Xformer \cite{zhang2023xformer}, Restormer \cite{zamir2022restormer}, PromptIR \cite{potlapalli2024promptir}, and adversarially robust AFM~\cite{ryou2024robust}, all of which come with open-source pre-trained models. 
For downstream tasks involving classification, we use a pre-trained ResNet50 \cite{he2016deep} as the classifier. In scenarios with unknown downstream tasks, we first train a MINE network as described in   \cref{subsec:mutual}. 
Since existing adversarial attacks on denoising models focus on degrading image clarity rather than explicitly manipulating semantics, we adopt the traditional adversarial attack method I-FGSM \cite{kurakin2018adversarial} as a representative baseline. The corresponding adversarial image is $x_{\text{adv}}$ and, after applying the denoising process, we get the denoised image $D(x_{\text{adv}})$.

\begin{figure}[t]
    \centering
    \includegraphics[width=0.48\textwidth]{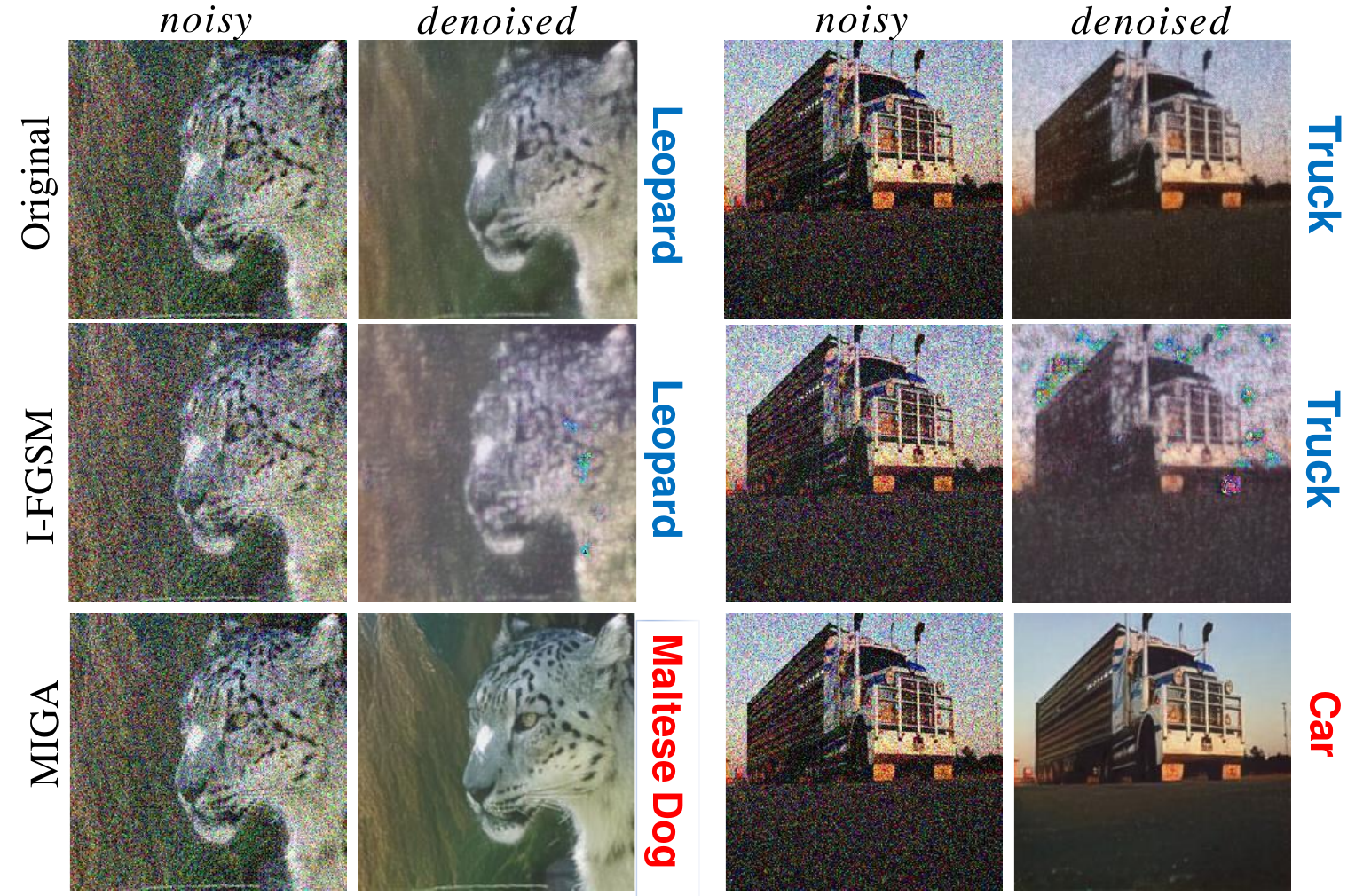} 
    \vspace{-0.5cm}
    \caption{Attack results on ImageNet-10. MIGA induces the denoising model to generate visually clean yet semantically altered outputs, leading pre-trained models to misclassify `Leopard' as `Maltese Dog' and `Truck' as `Car'.}

    \label{fig:attack-airline}
\end{figure}

\begin{table}[t]

    \vspace{-0.3cm}
    
    \centering
    \setlength{\tabcolsep}{2pt} 
    \renewcommand{\arraystretch}{0.9} 
    \resizebox{\columnwidth}{!}{ 
    \begin{tabular}{c|c|cccc|ccccc|ccccc}
        \toprule
        Network & Image & PSNR$\uparrow$ & SSIM$\uparrow$ & LPIPS$\downarrow$ & Entropy$\downarrow$ & Accuracy$\downarrow$ \\
        \midrule
        \multirow{2}{*}{\rotatebox{90}{}} 
        & $x$      & $\infty$ & 1.00 & 0.00 & 7.19 & 99.46 \\
        & $\xnoi$    & 17.65 & 0.30 & 0.62 & 7.71 & 83.00 \\
        \midrule
                \multirow{5}{*}{{Xformer}} 
        & $D(\xnoi)$ & 23.63 & 0.65 & 0.41 & 7.04 & 84.73 \\
        & $x_{\text{adv}}$ & 17.07 & 0.28 & 0.66 & 7.73 & 78.50 \\
        & $D(x_{\text{adv}})$ & 16.04 & 0.34 & 0.66 & 7.20 & 39.23 \\
        & $x_{\text{MIGA}}$ & 17.62 & 0.30 & 0.63 & 7.72 & 74.69 \\
        & $D(x_{\text{MIGA}})$ & 25.64 & 0.67 & 0.22 & 7.13 & 35.23 \\
        \midrule
        \multirow{5}{*}{{Restormer}} 
        & $D(\xnoi)$ & 22.36 & 0.70 & 0.38 & 6.81 & 89.85 \\
        & $x_{\text{adv}}$  & 17.21 & 0.29 & 0.65 & 7.72 & 76.85 \\
        & $D(x_{\text{adv}})$ & 15.98 & 0.57 & 0.56 & 6.72 & 64.77 \\
        & $x_{\text{MIGA}}$ & 17.60 & 0.30 & 0.63 & 7.72 & 70.62 \\
        & $D(x_{\text{MIGA}})$ & 30.77 & 0.78 & 0.26 & 7.06 & 51.54 \\
        \midrule

        \multirow{4}{*}{{PromptIR}}
        & $D(\xnoi)$ & 22.83 &0.50 & 0.45 & 7.56 & 85.77 \\
        & $x_{\text{adv}}$ & 16.46 & 0.26 & 0.67 & 7.77 & 64.65 \\
        & $D(x_{\text{adv}})$ & 11.94 & 0.11 & 0.80 & 7.78 & 34.77 \\
        & $x_{\text{MIGA}}$ & 17.61 & 0.30 & 0.63 & 7.72 & 74.46 \\
        & $D(x_{\text{MIGA}})$ & 24.20 & 0.59 & 0.37 & 7.60 & 21.31 \\

        \midrule

        \multirow{4}{*}{{AFM}}
        & $D(\xnoi)$ & 23.01 &0.71 & 0.42 & 7.01 & 86.43 \\
        & $x_{\text{adv}}$ & 17.21 & 0.28 & 0.64 & 7.72 & 79.12 \\
        & $D(x_{\text{adv}})$ & 16.09 & 0.58 & 0.61 & 6.92 & 70.27 \\
        & $x_{\text{MIGA}}$ & 19.32 & 0.32 & 0.62 & 7.63 & 74.92 \\
        & $D(x_{\text{MIGA}})$ & 31.28 & 0.78 & 0.21 & 7.03 & 42.99 \\
        \bottomrule
    \end{tabular}
    }
    \vspace{-0.3cm}
    \caption{Denoising attack for Imagenet-10.}
    \label{tab:main_kown}
   
\end{table}

\begin{table}[t]

    \centering
    \resizebox{\columnwidth}{!}{ 
    \begin{tabular}{c|c|cc}
    \toprule
    \multicolumn{1}{c|}{Metric}                      & \multicolumn{1}{c|}{Model} & \multicolumn{1}{c}{$ ImageNet-10_{\text{adv}}$} & \multicolumn{1}{c}{$ ImageNet-10_{\text{MIGA}}$} \\ 
    \midrule
    \multicolumn{1}{c|}{\multirow{3}{*}{$\text{LPIPS}_{\text{con}}$$\downarrow$}}
    & \multicolumn{1}{c|}{Restormer}  & 0.04 & 0.01 \\
    \multicolumn{1}{c|}{}                            & \multicolumn{1}{c|}{Xformer}  & 0.04  & 0.01 \\
    \multicolumn{1}{c|}{}                            & \multicolumn{1}{c|}
    {PromptIR} & 0.07  & 0.01 \\
    \multicolumn{1}{c|}{}                            & \multicolumn{1}{c|}
    {AFM} & 0.02  & 0.01 \\
 
    \bottomrule
    \end{tabular}
    }
\vspace{-0.3cm}
    \caption{Perturbation constraint for Imagenet-10.}
    \vspace{-0.4cm}
    \label{tab:constraint1}
\end{table}
\begin{figure*}[ht]
    \centering
    \includegraphics[width=0.85\textwidth]{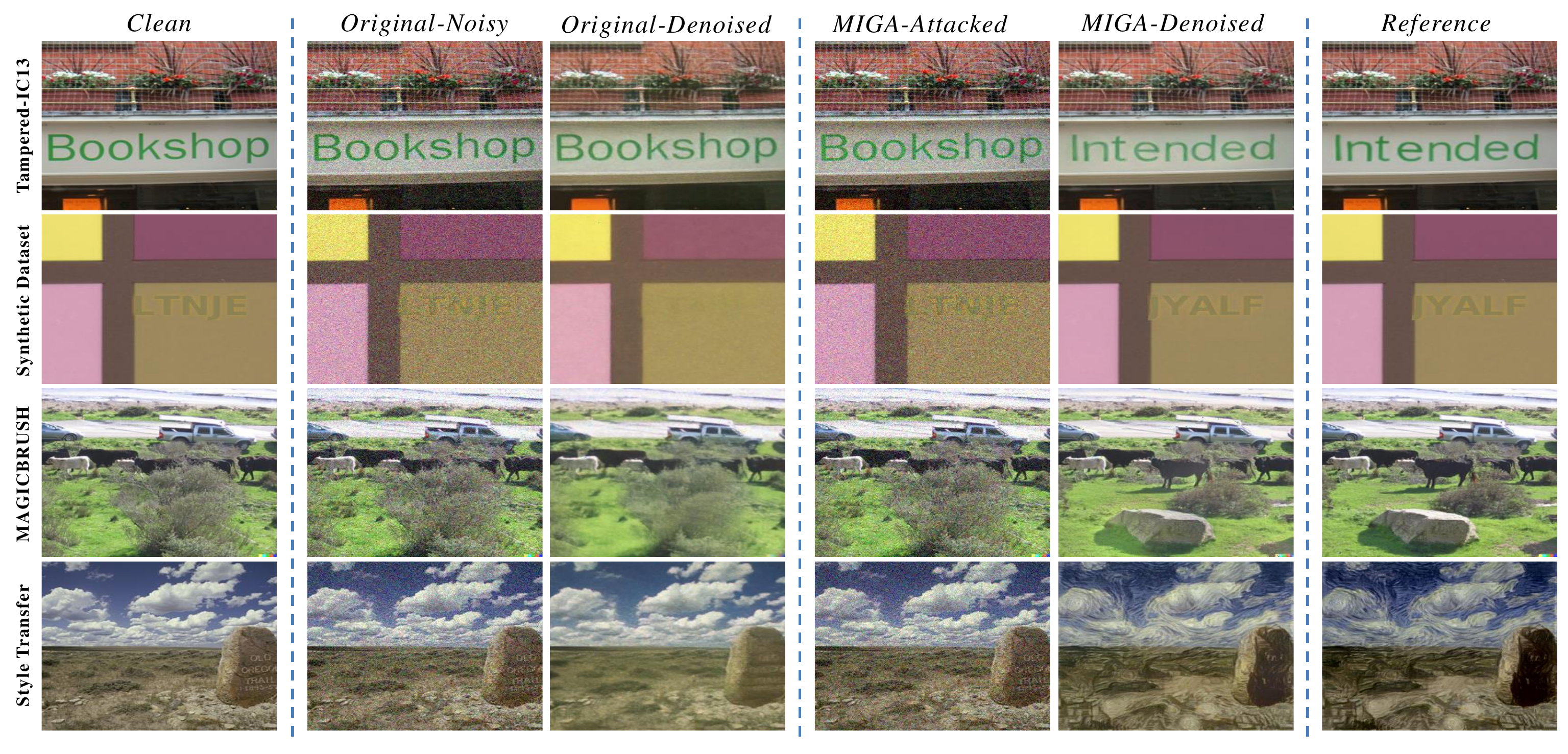} 
    \vspace{-0.5cm}
    \caption{{Attack results on unknown tasks. MIGA can induce the image denoising process to shift towards the semantic direction of the reference by adding small perturbations. Zoom in to see details better.}}
    \vspace{-0.4cm}

    \label{fig:main_unn}
\end{figure*}
\begin{table*}[ht]

    \centering
    \resizebox{\textwidth}{!}{ 
    \begin{tabular}{c|c|cccccc|cccc|cccc}
    
        \toprule
         \multirow{2}{*}{{Network}} & \multirow{2}{*}{Image} & \multicolumn{6}{c|}{Tampered-IC13} & \multicolumn{6}{c}{Synthetic Dataset} \\ 
        \cmidrule(lr){3-8} \cmidrule(lr){9-14}
         &  & PSNR$\uparrow$ & SSIM$\uparrow$ & LPIPS$\downarrow$ & Entropy$\downarrow$ & ROUGE-L$\uparrow$ &  & PSNR$\uparrow$ & SSIM$\uparrow$ & LPIPS$\downarrow$ & Entropy$\downarrow$ & ROUGE-L$\uparrow$ &  \\ 
 
        \midrule

        \multirow{5}{*}{{Xformer}} 
         & $x$    & $\infty$ & 1.00 & 0.00 & 6.60 & 0.57 & & $\infty$ & 1.00 & 0.00 & 5.05 & 0.15 &  \\
        & $\xnoi$  & 23.98 & 0.47 & 0.48 & 7.32 & --   & & 24.15 & 0.28 & 0.65 & 6.61 & --   &  \\
        & $D(\xnoi)$   & 27.57 & 0.85 & 0.30 & 6.61 & 0.57 & & 31.70 & 0.90 & 0.39 & 5.10 & 0.15 &  \\
        & $x_{\text{MIGA}}$ & 22.92 & 0.45 & 0.48 & 7.34 & --   & & 24.07 & 0.28 & 0.65 & 6.61 & --   &  \\
        & $D(x_{\text{MIGA}})$ & 25.91 & 0.86 & 0.15 & 6.74 & 0.83 & & 34.84 & 0.94 & 0.14 & 5.05 & 0.75 &  \\
        \hline

    \end{tabular}
    }
\vspace{-0.3cm}
        \caption{Denoising Attack for Text Alteration Tasks.}
    
    \vspace{-0.3cm}
    \label{tab:Alteration}
\end{table*}

\begin{table*}[ht!]

    \centering
    \resizebox{\textwidth}{!}{ 
    \begin{tabular}{c|c|cccccc|cccccc}
        \toprule
        \multirow{2}{*}{{Network}} & \multirow{2}{*}{Image} & \multicolumn{6}{c|}{MAGICBRUSH} & \multicolumn{6}{c}{Style Transfer} \\ 
        \cmidrule(lr){3-8} \cmidrule(lr){9-14}
         &  & PSNR$\uparrow$ & SSIM$\uparrow$ & LPIPS$\downarrow$ & Entropy$\downarrow$ & CLIP similarity$\uparrow$ &  & PSNR$\uparrow$ & SSIM$\uparrow$ & LPIPS$\downarrow$ & Entropy$\downarrow$ & Gram matrix Loss$\downarrow$ &  \\

        \midrule
        \multirow{5}{*}{{Xformer}} 
         
        & $x$       & $\infty$ & 1.00 & 0.00 & 7.36  & 0.23 &  & $\infty$ & 1.00 & 0.00 & 7.15  & 74.58   \\ 
        & $\xnoi$ & 34.56 & 0.94 & 0.07 & 7.40  & --   &  & 24.09 & 0.57 & 0.45 & 7.51  & --   & \\ 
        & $D(\xnoi)$ & 36.60 & 0.97 & 0.05 & 7.36  & 0.23 &  & 25.56 & 0.79 & 0.34 & 7.03  & 23.75  & \\ 
        & $x_{\text{MIGA}}$ & 31.14 & 0.92 & 0.08 & 7.40  & --   &  & 21.76 & 0.52 & 0.45 & 7.52  & --   & \\ 
        & $D(x_{\text{MIGA}})$ & 26.75 & 0.94 & 0.05 & 7.37  & 0.24 &  & 21.00 & 0.55 & 0.21 & 7.22  & 9.10  & \\ 
        
        \bottomrule
    \end{tabular}}
    
 \vspace{-0.3cm}
    \caption{Denoising Attack for content replacement and style transfer tasks.}
   \label{tab:replacement}
    \vspace{-0.5cm}
\end{table*}

\begin{table}[t]

    \centering
    \resizebox{\columnwidth}{!}{ 
    \begin{tabular}{c|cccc}
    \toprule
    \multicolumn{1}{c|}{Metric} & \multicolumn{1}{l}{Tampered-IC13} & \multicolumn{1}{l}{Synthetic Dataset} & \multicolumn{1}{l}{MAGICBRUSH} & \multicolumn{1}{l}{Style Transfer} \\ 
    \midrule

    \text{$\text{LPIPS}_{\text{con}}$$\downarrow$} & 0.02  & 0.01  & 0.01  & 0.04  \\ 
    \bottomrule
    \end{tabular}
    }
    \vspace{-0.3cm}
    \caption{Perturbation constraint for unknown task.}
    
    \vspace{-0.7cm}
    \label{tab:constraint7}
\end{table}

\subsection{Attack denoising models with known tasks }

In this subsection, we evaluate the effectiveness of our proposed method, MIGA, to attack denoising models in scenarios where the downstream task is known, focusing on image classification using the ImageNet-10 dataset. We compare MIGA with I-FGSM, assessing both  denoised image clarity and classification accuracy.  ~\cref{tab:main_kown} summarizes the results, where higher PSNR and SSIM values, along with lower LPIPS and entropy, indicate better image clarity, and a reduction in classification accuracy reflects semantic alteration.
The key findings are as follows:

\begin{itemize}
    \item \textbf{Improvement in downstream task performance by denoising:} The denoising process using $D$ improves downstream task performance. For all four denoising models the accuracy of the noisy image $\xnoi$ increases significantly after denoising, reaching 84.73\%, 89.85\%, 85.77\%, and 86.43\%, respectively.
    
\item \textbf{Limitations of traditional attacks:}  
I-FGSM aims to reduce the clarity of denoised images. However, this reduction comes at the cost of significant image quality degradation. As shown in ~\cref{fig:attack-airline}, I-FGSM introduces color distortions and artifacts, causing a noticeable loss in image quality. Despite this degradation, both `Truck' and `Leopard' remain correctly classified.

    \item \textbf{Effectiveness of MIGA:} MIGA consistently enhances denoised image clarity while notably reducing classification accuracy (see  ~\cref{fig:attack-airline} and  ~\cref{tab:main_kown}). For example, in the Restormer model, MIGA achieves a PSNR of 30.77 and a low LPIPS of 0.26 after denoising $D(x_{\text{MIGA}})$, indicating high image quality. Meanwhile, the classification accuracy drops to 51.54\%, demonstrating the effectiveness of our method in altering semantic information during the denoising process without degrading image quality.
    
    \item \textbf{Imperceptible perturbations:} MIGA introduces minimal perturbations that result in negligible perceptual differences, as shown in  ~\cref{tab:constraint1}. The low LPIPS values (e.g., 0.01 for Restormer) ensure that the attack remains undetected while affecting the downstream task.
\end{itemize}

These results show that MIGA effectively performs semantic attacks on denoising models in scenarios where the downstream task is known, achieving high-quality denoising while significantly reducing downstream task performance without introducing noticeable artifacts.

\subsection{Attack denoising models with unknown tasks}
\label{main_un_results}

We evaluate our method's performance on denoising models across tasks with unknown downstream objectives, including text alteration, content replacement, and style transfer, using corresponding reference images. Some results are visualized in   \cref{fig:main_unn}. The semantic metrics used are ROUGE-L for text similarity, CLIP for content editing, and Gram matrix loss for style transfer.

\textbf{Results on text alteration tasks.} 
 ~\cref{tab:Alteration} presents the results for the Tampered-IC13 and Synthetic datasets. We observe that even without knowledge of the downstream task, MIGA can successfully alter the semantic content after denoising while maintaining high image quality. For instance, in the Xformer model on the Tampered-IC13 dataset, the denoised image $D(x_{\text{MIGA}})$ achieves a high PSNR of 25.91 and a low LPIPS of 0.15, while the ROUGE-L score increases to 0.83, indicating significant text alteration.

\textbf{Results on content replacement and style transfer.} 
 ~\cref{tab:replacement} shows the results for content replacement using MAGICBRUSH and style transfer tasks. Our method effectively performs semantic modifications in these tasks as well. For example, in the style transfer task, the Gram matrix loss decreases to 9.10 after denoising $D(x_{\text{MIGA}})$, indicating successful style transfer, while maintaining acceptable image quality (Entropy  of 7.22).

\textbf{Minimal perturbations.} 
As shown in ~\cref{tab:constraint7}, the perturbations introduced are minimal, with very low LPIPS (e.g., LPIPS of 0.01 for the Synthetic dataset). This confirms that our method introduces imperceptible changes to the images while disrupting the denoising performance.

\subsection{Ablation study}
We perform an ablation study to assess the contributions of different loss components in our method.

\textbf{Importance of loss components.}
 ~\cref{unknown weight} summarizes the impact of removing or modifying various loss terms on the Tampered-IC13 dataset using the Xformer model. The key observations are:

\begin{itemize} \item \textbf{Mutual information loss ($\mathcal{L}_{\text{MI}}$):} Removing $\mathcal{L}_{\text{MI}}$ leads to a decrease in the ROUGE-L score from 0.83 to 0.73, indicating less effective semantic alteration. This highlights the importance of $\mathcal{L}_{\text{MI}}$ in influencing semantic alteration during attacking denoising models. \item \textbf{Reconstruction loss ($\mathcal{L}_{\text{rec}}$):} Omitting $\mathcal{L}_{\text{rec}}$ increases LPIPS to 0.40, indicating a significant decline in image quality after denoising. This shows that $\mathcal{L}_{\text{rec}}$ is crucial for maintaining denoised image clarity. \item \textbf{Perturbation constraint loss ($\mathcal{L}_{\text{com}}$):} Removing $\mathcal{L}_{\text{com}}$ results in the highest ROUGE-L score (0.92) but introduces noticeable perturbations ({$\text{LPIPS}_{\text{con}}$} increases to 0.38), compromising the imperceptibility of the attack. \item \textbf{Perceptual loss and MSE Loss:} Using only MSE loss or only perceptual loss adversely affects both image clarity and the  effectiveness of semantic alteration. This demonstrates that the combination of loss terms is necessary for optimal performance. \end{itemize}
\begin{table}[t]

    \centering
    \setlength{\tabcolsep}{2pt} 
    \renewcommand{\arraystretch}{0.9} 
    \resizebox{\columnwidth}{!}{ 
    \begin{tabular}{cc|cc|c|cccc}
        \toprule
        \multicolumn{2}{c|}{$L {\text{con}}$} & \multicolumn{2}{c|}{$L {\text{rec}}$} & \multirow{2}{*}{$L_{\text{MI}}$} & \multicolumn{4}{c}{Metrics} \\
        \cmidrule(lr){1-2} \cmidrule(lr){3-4} \cmidrule(lr){6-9}
        MSE & $L {\text{perc}}$ & MSE & $L {\text{perc}}$ &  & $\text{LPIPS}_{\text{con}}$$\downarrow$  & LPIPS$\downarrow$  & Entropy$\downarrow$  & ROUGE-L$\uparrow$ \\
        \midrule
        \checkmark & \checkmark & \checkmark & \checkmark & \checkmark & 0.02  & 0.15 & 6.74 & 0.83 \\
        \checkmark & \checkmark & \checkmark & \checkmark & $\times$ & 0.02  & 0.18 & 6.74 & 0.73 \\
        \checkmark & \checkmark & $\times$ & $\times$ & \checkmark & 0.01  & 0.40 & 6.80 & 0.50 \\
        $\times$ & $\times$ & \checkmark & \checkmark & \checkmark & 0.38  & 0.12 & 6.67 & 0.92 \\
        \checkmark & $\times$ & \checkmark & $\times$ & \checkmark & 0.08  & 0.40 & 6.87 & 0.48 \\
        $\times$ & \checkmark & $\times$ & \checkmark & \checkmark & 0.03  & 0.19 & 6.76 & 0.76 \\
        \bottomrule
    \end{tabular}}
\vspace{-0.3cm}
    \caption{Importance of different losses for Tampered-IC13 dataset.}
    \label{unknown weight}
    \vspace{-0.3cm}
\end{table}

\begin{table}[t]

    \centering
    \resizebox{\columnwidth}{!}{ 
    \begin{tabular}{c|c|cccc|c}
        \toprule
        Difficulty & $\text{LPIPS}_{\text{con}}$ & PSNR$\uparrow$ & SSIM$\uparrow$ & LPIPS $\downarrow$ & Entropy$\downarrow$  &ROUGE-L$\uparrow$ \\
        \midrule
        Easy & 0.01 & 34.62 & 0.94 & 0.17 & 5.00 & 0.54 \\
        Medium & 0.01 & 34.09 & 0.94 & 0.17 & 5.05 & 0.69 \\
        Difficult & 0.01 & 32.55 & 0.93 & 0.19 & 5.21 & 0.73 \\
        \bottomrule
    \end{tabular}}
\vspace{-0.3cm}
    \caption{Difficulty evaluation.}
    \label{tab:diff}
    
\end{table}
\textbf{Effect of task difficulty.} We   examine the impact of task difficulty by varying the font size in the Synthetic dataset. As shown in ~\cref{tab:diff}, as the difficulty increases (larger font sizes), there is a slight sacrifice in image clarity (LPIPS increases), but the ROUGE-L score also increases, indicating more effective semantic alteration. Nonetheless, the perturbations remain imperceptible ($\text{LPIPS}_{\text{con}}$ remains at 0.01).

\subsection{Robustness of adversarial examples}

\begin{figure}[t]
    \centering
    \includegraphics[width=0.5\textwidth]{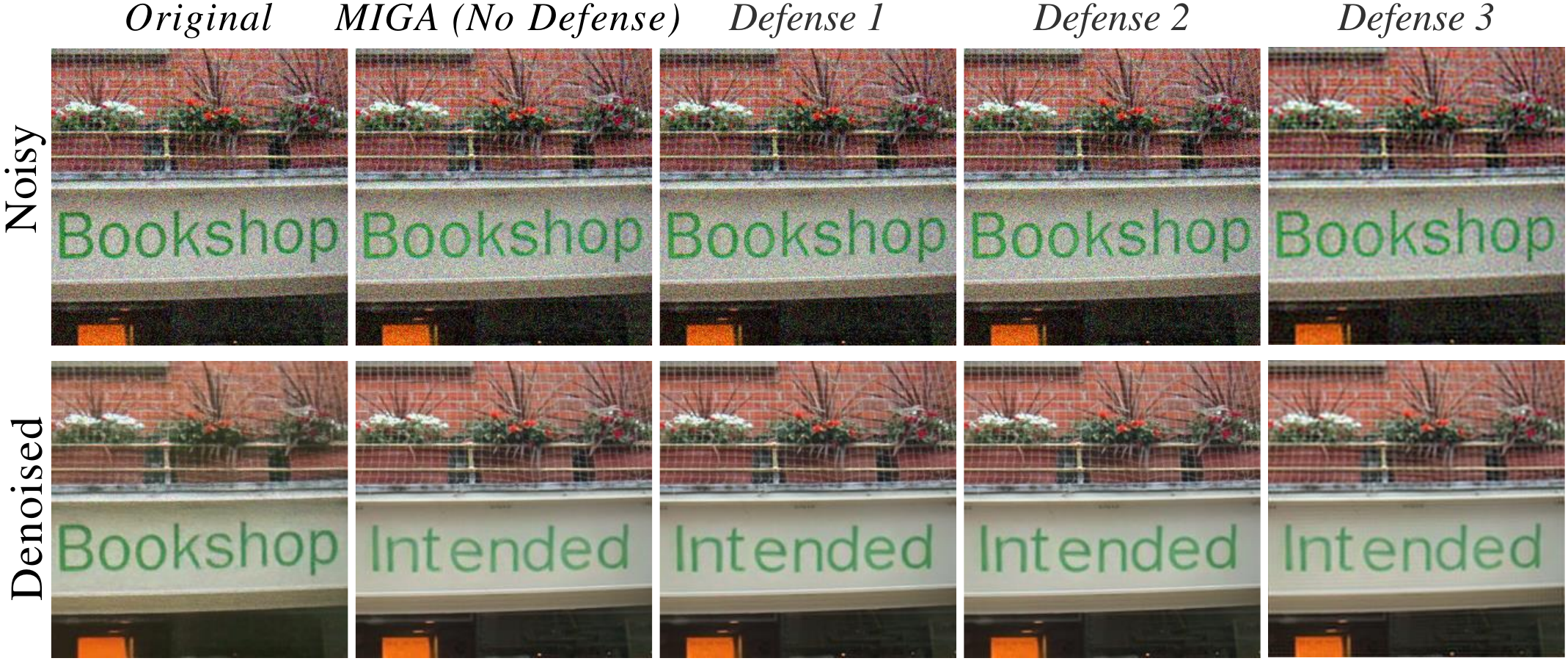} 
  
    \caption{{Impact of different defense strategies on MIGA. A comparison of feature squeezing (Defense 1), non-local means denoising (Defense 2), and DiffPure (Defense 3) on image denoising performance.}}

    \label{fig:roub}
\end{figure}

We assess the robustness of our adversarial examples against several common defense strategies, including Feature Squeezing \cite{xu2017feature}, Non-Local Means Denoising \cite{buades2005non}, and Randomized Smoothing \cite{chiang2020certified}. As shown in  ~\cref{tab:robustness} and  ~\cref{fig:roub}, the semantic modifications introduced by our method remain detectable under these defense conditions, with visible alterations in all cases. Notably, these techniques do not significantly degrade image quality, as indicated by stable PSNR and LPIPS values. Moreover, the semantic changes persist despite the application of defenses, as reflected by the high ROUGE-L scores (e.g., 0.80 under Feature Squeezing), which demonstrate the resilience of our adversarial examples to these countermeasures. These results highlight the effectiveness and robustness of our attack against widely used defense strategies.

\begin{table}[t]

    \centering
    \resizebox{\columnwidth}{!}{ 
    \begin{tabular}{c|cc c c| c}
        \hline
        Defense   &PSNR$\uparrow$&SSIM$\uparrow$& LPIPS$\downarrow$ & Entropy$\downarrow$ & ROUGE-L$\uparrow$ \\
        \midrule
        Origin   & 23.98  & 0.47 & 0.48 & 7.32  & --    \\
 
        No-Defense  & 25.91 & 0.86 & 0.15 & 6.74 & 0.83 \\
        Feature Squeezing & 26.04 & 0.87 & 0.13 & 6.73 & 0.80 \\
        Non Local Means Denoising  & 25.53 & 0.87 & 0.13 & 6.74 & 0.80 \\
        DiffPure  & 25.92 & 0.85 & 0.16 & 6.73 & 0.77 \\
        \hline
    \end{tabular}}

    \caption{Robustness evaluation}
    \label{tab:robustness}
    \vspace{-0.5cm}
\end{table}

\section{ Conclusion}

In this work, we propose MIGA, an adversarial attack framework that directly targets image denoising models by disrupting their ability to preserve semantic information. By leveraging mutual information (MI) to quantify semantic consistency, MIGA introduces imperceptible perturbations that selectively alter task-relevant semantics while maintaining visual fidelity. We design three tailored loss functions to optimize this process, enabling attacks in both known and unknown downstream task scenarios. Extensive experiments across multiple denoising models validate MIGA’s effectiveness and robustness, revealing a critical security risk of denoising models in real-world applications. This work also complements existing adversarial attacks and provides insights into enhancing the robustness of denoising models against semantic manipulation.

\clearpage
\clearpage
\setcounter{page}{1}

\appendix

\section{Proof of task-relevant mutual information reduction} \label{sec:proof}

As introduced in Sec.   4.1, our definition of task-relevant mutual information focuses on the semantic features directly relevant to the downstream task, differing from the traditional information-theoretic definition. In this section, we rigorously prove that, when the downstream task model \( F \) is known, maximizing the cross-entropy loss \( \mathcal{L}_{\text{CE}}(F(D(\xnoi + \delta)), C) \) is equivalent to minimizing the task-relevant mutual information \( I(x; D(\xnoi + \delta) \mid C) \) between the original image \( x \) and the denoised image \( D(\xnoi + \delta) \), conditioned on the ground-truth label \( C \).

\begin{lemma}
Increasing the cross-entropy loss \( \mathcal{L}_{\text{CE}}(F(D(\xnoi + \delta)), C) \) reduces the task-relevant mutual information \( I(x; D(\xnoi + \delta) \mid C) \) between the original and denoised images, conditioned on \( C \).
\end{lemma}

\begin{proof}
We begin by clarifying the notation:
\begin{itemize}
    \item \( x \) denotes the original image.
    \item \( \xnoi \) represents the noisy version of \( x \).
    \item \( D(\xnoi + \delta) \) is the denoised image after applying a perturbation \( \delta \).
    \item \( C \) is the random variable representing the ground-truth class label corresponding to \( x \).
    \item \( c \) is a particular class label in the set of all possible classes \( \mathcal{C} \).
    \item \( F \) is the downstream classification model, estimating the probability \( P_F(c \mid D(\xnoi + \delta)) \) that the denoised image \( D(\xnoi + \delta) \) belongs to class \( c \).
\end{itemize}

Recall the definition of task-relevant mutual information:
\begin{equation} \label{eq:mutual_info}
I(x; D(\xnoi + \delta) \mid C) = H(C \mid x) - H(C \mid D(\xnoi + \delta)),
\end{equation}
where \( H(C \mid x) \) and \( H(C \mid D(\xnoi + \delta)) \) are the conditional entropies of the label \( C \) given the images \( x \) and \( D(\xnoi + \delta) \), respectively.

Since \( x \) uniquely determines \( C \) ( \( x \) perfectly predicts \( C \)), we have \( H(C \mid x) = 0 \). Thus, the task-relevant mutual information simplifies to:
\begin{equation} \label{eq:mutual_info_simplified}
I(x; D(\xnoi + \delta) \mid C) = - H(C \mid D(\xnoi + \delta)).
\end{equation}
Our goal is to minimize \( I(x; D(\xnoi + \delta) \mid C) \), which is equivalent to maximizing \( H(C \mid D(\xnoi + \delta)) \).

The conditional entropy \( H(C \mid D(\xnoi + \delta)) \) is defined as:
\begin{equation} \label{eq:conditional_entropy}\begin{aligned}
&H(C \mid D(\xnoi + \delta)) = \\
&\quad - \mathbb{E}_{D(\xnoi + \delta)} \left[ \sum_{c \in \mathcal{C}} P(c \mid D(\xnoi + \delta)) \log P(c \mid D(\xnoi + \delta)) \right],
\end{aligned}
\end{equation}
where \( P(c \mid D(\xnoi + \delta)) \) is the true probability of class \( c \) given the image \( D(\xnoi + \delta) \). The expectation \( \mathbb{E}_{D(\xnoi + \delta)} \) is taken over the distribution of \( D(\xnoi + \delta) \), which represents the probabilistic outputs of the denoising model \( D \) given the perturbed input \( \xnoi + \delta \). 

In practice, the true conditional probability \( P(c \mid D(\xnoi + \delta)) \) is unknown. Instead, we use the high-accuracy and well-calibrated classifier \( F \) to estimate it:
\begin{equation} \label{eq:estimated_prob}
{P(c \mid D(\xnoi + \delta)) \approx P_F(c \mid D(\xnoi + \delta)),}
\end{equation}
where 
$P(c \mid D(\xnoi + \delta))$is the probability output by the classifier 
 \( F \) for class  \( c \).The approximation's accuracy depends on 
the  performance of \( F \), which we assume is sufficiently high for this analysis.
 
Substituting this into Equation~\eqref{eq:conditional_entropy}, we approximate the conditional entropy:

\begin{multline} \label{eq:approx_conditional_entropy}
H(C \mid D(\xnoi + \delta)) \approx \\
\quad - \mathbb{E}_{D(\xnoi + \delta)} \Bigg[ \sum_{c \in \mathcal{C}} P_F(c \mid D(\xnoi + \delta))\log P_F(c \mid D(\xnoi + \delta)) \Bigg].
\end{multline}

Recall that the cross-entropy loss between the classifier's prediction and the ground-truth label \( C \) is:
\begin{equation} \label{eq:cross_entropy_loss}
\mathcal{L}_{\text{CE}}(F(D(\xnoi + \delta)), C) = - \log P_F(C \mid D(\xnoi + \delta)).
\end{equation}
Increasing \( \mathcal{L}_{\text{CE}} \) is equivalent to reducing \( P_F(C \mid D(\xnoi + \delta)) \), i.e., decreasing the classifier's confidence in the correct class \( C \). Due to the normalization condition
\begin{equation} \label{eq:prob_normalization}
\sum_{c \in \mathcal{C}} P_F(c \mid D(\xnoi + \delta)) = 1.
\end{equation}
A decrease in \( P_F(C \mid D(\xnoi + \delta)) \) leads to an increase in \( P_F(c \mid D(\xnoi + \delta)) \) for \( c \neq C \). Assuming no bias or preference for any specific incorrect class, this redistribution results in a more uniform probability distribution \( P_F(c \mid D(\xnoi + \delta)) \) closer to the uniform distribution. Consequently, as \( P_F(c \mid D(\xnoi + \delta)) \) becomes more uniform, the conditional entropy \( H(C \mid D(\xnoi + \delta)) \) increases, consistent with Shannon's entropy theorem~\cite{shannon1948mathematical}.

Therefore, maximizing \( \mathcal{L}_{\text{CE}} \) effectively increases \( H(C \mid D(\xnoi + \delta)) \). Substituting back into Equation~\eqref{eq:mutual_info_simplified}, we find that \( I(x; D(\xnoi + \delta) \mid C) \) decreases. Thus, maximizing the cross-entropy loss \( \mathcal{L}_{\text{CE}} \) minimizes the task-relevant mutual information \( I(x; D(\xnoi + \delta) \mid C) \).
\end{proof}

\section{Task-Relevant mutual information for unknown downstream tasks}\label{sec:unknown_task_mi}

In the main text, we define the task-relevant mutual information 
$I(x; D(\xnoi + \delta) \mid C)$ 
to quantify the shared information between the original image $x$ and the denoised image $D(\xnoi + \delta)$ that is relevant to a downstream task with ground-truth output $C$.

When the downstream task $F$ is unknown and $C$ is unobservable, directly computing $I(x; D(\xnoi + \delta) \mid C)$ becomes infeasible. To address this challenge, we propose an approximation strategy that leverages semantically modified images $x_{\text{alt}}$.

Our key assumption is that the semantic alterations introduced in $x_{\text{alt}}$ capture features relevant to potential downstream tasks. By carefully modifying key semantic regions of the original images $x$, we simulate changes that are likely to affect any reasonable downstream task output.

Under this assumption, we approximate the task-relevant mutual information by focusing on the mutual information between $x$ and $D(\xnoi + \delta)$ with an emphasis on task-relevant features. We achieve this by training a MINE network $T$ \cite{belghazi2018mine} using contrastive learning:

\begin{itemize}
    \item \textbf{Positive pairs:} $(x, D(\xnoi))$, where the denoised image retains the original semantics.
    \item \textbf{Negative pairs:} $(x, x_{\text{alt}})$, where the semantics have been altered.
\end{itemize}

By maximizing similarity for positive pairs and minimizing it for negative pairs, $T$ learns to focus on task-relevant features. During the attack, we optimize the adversarial perturbation $\delta$ to minimize $T(x, D(\xnoi + \delta))$, effectively reducing the estimated mutual information related to these features.

This approach allows us to approximate the minimization of $I(x; D(\xnoi + \delta) \mid C)$ without direct access to $C$. By relying on semantically altered images as proxies for task-relevant changes, we ensure that our MIGA remains effective even when the downstream task is unknown, aligning with our original formulation.

\section{Experimental settings} 
\label{sec:details}
\subsection{Datasets} \label{sec:Datasets}
In this appendix, we provide detailed descriptions of the datasets used in our experiments, focusing on their specific usage and processing steps.

\paragraph{1.ImageNet-10.}  
The ImageNet-10 dataset \cite{russakovsky2015imagenet} consists of images from 10 distinct classes. We use this dataset to evaluate classification tasks under noisy conditions. To simulate noise, we add Gaussian noise with a standard deviation \(\sigma = 50\) to each image. The images retain their original sizes as provided in the dataset.

\paragraph{2.Tampered-IC13 dataset.}  
The Tampered-IC13 dataset \cite{wang2022detecting} contains images with real-world text alterations in natural scenes, making it suitable for evaluating text tampering detection methods. We apply OCR to detect and recognize text regions within each image. The original images vary in size. For each detected text region, we extract a crop of size \(256 \times 256\) pixels centered on the text location to standardize the input size. We then add Gaussian noise with \(\sigma = 25\) to each cropped image, resulting in a total of 404 noisy images.

\paragraph{3.MAGICBRUSH dataset.}  
The MAGICBRUSH dataset \cite{zhang2024magicbrush} is a large-scale, manually annotated, instruction-guided image editing dataset. It covers diverse scenarios, including single-turn and multi-turn edits, as well as mask-provided and mask-free editing tasks. The dataset contains 10,000 (source image, instruction, target image) triples, making it ideal for training and evaluating image editing models. We resize all original images, which vary in size, to \(256 \times 256\) pixels to ensure consistency. Gaussian noise with \(\sigma = 25\) is added to the source images.

\paragraph{4.Neural style Transfer dataset.}  
This synthetic dataset is designed to evaluate models on stylized images. We select images from the DFWB datasets (DIV2K~\cite{agustsson2017ntire}, Flickr2K, WED~\cite{ma2016waterloo}, BSD~\cite{martin2001database}) and resize them to \(256 \times 256\) pixels. Gaussian noise with \(\sigma = 25\) is added to these images to serve as noisy inputs. We use the neural style transfer method from {Gatys et al.~\cite{gatys2015neural}}, training the style transfer model for 500 epochs to generate stylized versions of the images. These stylized images serve as the target images, resulting in a total of 2,921 image pairs.

\paragraph{5.Synthetic text Alteration dataset.}  
We create this dataset to evaluate models on text alteration tasks involving small but semantically significant changes. We select 3,000 patches from the SIDD dataset \cite{abdelhamed2018high}, each resized to \(256 \times 256\) pixels. Two sets of text overlays, containing five different digits and letters, are placed at the same location within each image patch. One set is corrupted with Gaussian noise (\(\sigma = 25\)) to serve as the noisy original images, while the other set remains clean to serve as the target images.

\vspace{1em}

\begin{figure*}[ht]
    \centering
    \includegraphics[width=0.85\textwidth]{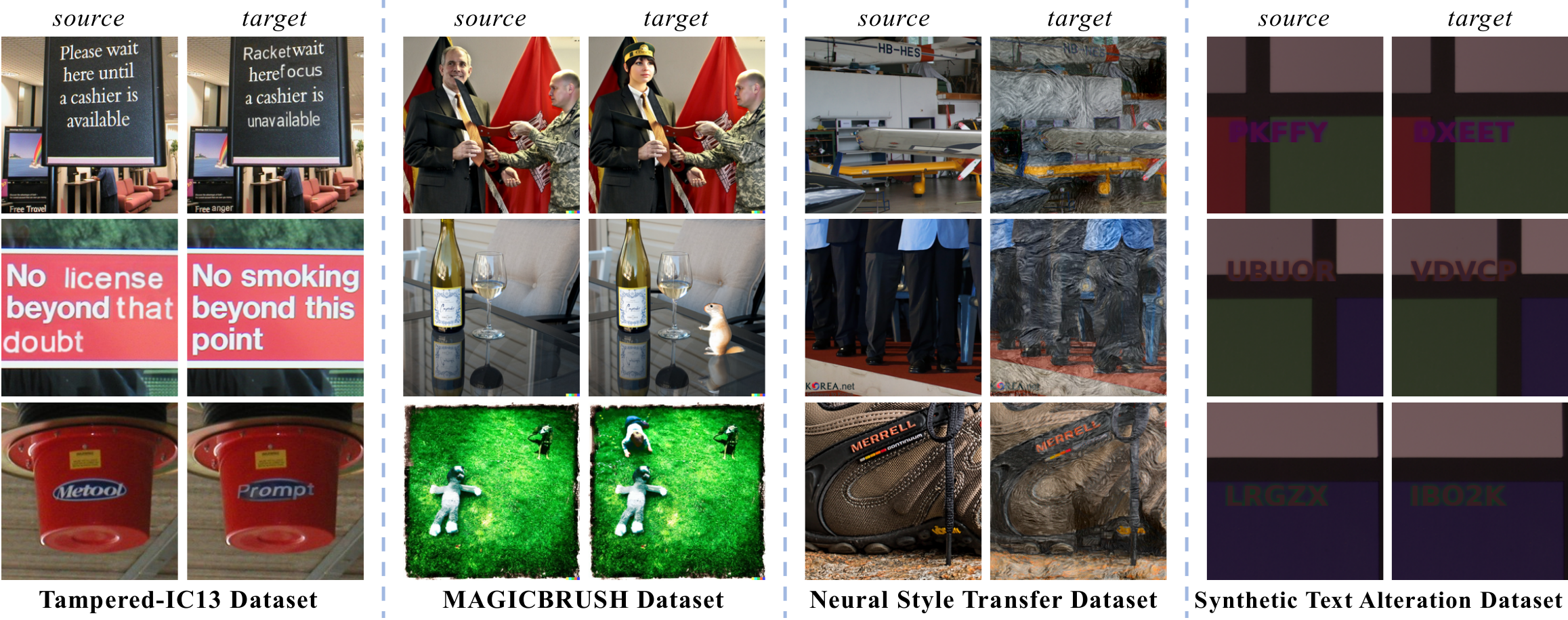} 
    \caption{Example images from our datasets, including a pair of related images with differing semantics.}

    \label{fig:dd1}
    \vspace{-0.4cm}
\end{figure*}

In our experiments, we add Gaussian noise to images across different datasets to create noisy versions. Specifically, Gaussian noise with \(\sigma = 50\) is applied to the ImageNet-10 dataset, while Gaussian noise with \(\sigma = 25\) is used for all other datasets. When computing the reconstruction loss \(\mathcal{L}_{\text{rec}}\), we use the clean image \(x_{\text{ref}}\) as the reference without any added noise.
Some visualization results of the clean aforementioned dataset are shown in \cref{fig:dd1}.

\subsection{Evaluation metrics}\label{sec:Metrics}

To thoroughly evaluate the performance of our algorithm, we employ a comprehensive set of metrics focusing on three key aspects: imperceptibility of the perturbations, clarity of the denoised images, and semantic fidelity with respect to downstream tasks.

\subsubsection{Imperceptibility measures}

To assess the imperceptibility of the adversarial perturbations, we compute the following metrics between the noisy image \( \xnoi \) and the adversarial image \( \xnoi + \delta \):

\begin{itemize}
    \item \textbf{PSNR\(_{\text{con}}\):} It measures the pixel-wise similarity, with higher values indicating smaller differences and thus less perceptible perturbations \cite{wang2002universal}.
    
    \item \textbf{SSIM\(_{\text{con}}\):} It evaluates structural similarity, with values closer to 1 indicating that the perturbations are structurally imperceptible \cite{wang2002universal}.
    
    \item \textbf{LPIPS\(_{\text{con}}\):} It quantifies perceptual similarity, with lower scores indicating higher perceptual similarity and thus more imperceptible perturbations \cite{zhang2018unreasonable}.
    
    \item \textbf{Entropy\(_{\text{con}}\):} It measures the difference in image entropy between \( \xnoi \) and \( \xnoi + \delta \). Minimal changes in entropy indicate that the perturbations do not introduce noticeable randomness or complexity, helping to keep the perturbations imperceptible \cite{shannon1948mathematical}.
\end{itemize}

These imperceptibility metrics, denoted with the subscript \(_{\text{con}}\), ensure that the adversarial perturbations \( \delta \) added to the noisy images are imperceptible to human observers, thus maintaining the stealthiness of the attack.

\subsubsection{Distortion and clarity measures}

To evaluate the clarity and quality of the denoised images, we compute the following metrics between the denoised images and the original clean images:

\begin{itemize}
    \item \textbf{PSNR:} It measures the pixel-wise similarity between the original image \( x \) and the denoised images. Higher PSNR values indicate better denoising quality \cite{wang2002universal}.
    
    \item \textbf{SSIM:} It assesses the structural similarity between the original and denoised images, with values closer to 1 indicating higher structural fidelity \cite{wang2002universal}.
    
    \item \textbf{LPIPS:} It computes perceptual similarity between the original image \( x \) and the denoised images, with lower scores indicating better perceptual quality \cite{zhang2018unreasonable}.
    
    \item \textbf{Entropy:} It evaluates the amount of information or randomness in the denoised images. Appropriate levels of entropy indicate the preservation of image details without introducing artifacts. An excessively high entropy may indicate residual noise or artifacts, while  an entropy value that is too low may suggest over-smoothing and loss of important details. \cite{shannon1948mathematical}.
\end{itemize}

These metrics are calculated using clean reference images, $x_{\text{ref}}$, as the benchmark. Our objective is to ensure that the denoised outputs, $D(\xnoi)$ and $D(\xnoi + \delta)$, exhibit high quality and maintain a close resemblance to the corresponding clean images. This demonstrates the efficacy of the denoising process, even when faced with adversarial perturbations.

\subsubsection{Semantic fidelity measures}

To assess the impact of adversarial perturbations on semantic alterations, we employ task-specific performance metrics of downstream tasks, providing detailed evaluations for each scenario.

\begin{itemize}
\item \textbf{Classification tasks:}  
For evaluating classification tasks, we use a pretrained downstream classifier \( F \) to classify the denoised adversarial images \( D(\xnoi + \delta) \). We measure the classification accuracy and compare it with that of the clean images \( x \) and the denoised images \( D(\xnoi) \). A significant decrease in accuracy on \( D(\xnoi + \delta) \) indicates that the adversarial perturbations have successfully altered the semantic content of the denoised images, causing the classifier to misclassify them. This demonstrates the effectiveness of the attack in compromising the semantic fidelity of denoising models.

    \item \textbf{Text alteration:} In tasks involving text recognition, such as Optical Character Recognition (OCR) \cite{singh2012survey}, we employ the {ROUGE-L} metric \cite{lin2004rouge} to evaluate the semantic changes introduced by adversarial perturbations. This metric measures the Longest Common Subsequence (LCS) between the original text and the OCR results obtained from the reference image (which has been semantically modified), capturing differences in lexical choices and sentence structure. A higher ROUGE-L score indicates that the adversarial perturbations have successfully caused significant semantic deviations, demonstrating their effectiveness in altering the textual content during denoising.

\item \textbf{Image modification:} For tasks where images are supposed to be modified according to textual instructions (e.g., adding an object, changing colors), we evaluate how well the denoised adversarial image \( D(\xnoi + \delta) \) aligns with the intended modification instructions. We employ Contrastive Language–Image Pre-training {(CLIP)} similarity \cite{radford2021learning}, which computes the cosine similarity between the image and text embeddings in a shared multimodal space. CLIP has been shown to effectively capture semantic relationships between images and text. A higher CLIP similarity score between \( D(\xnoi + \delta) \) and the modification instructions indicates that the adversarial perturbations have successfully enabled the denoised image to adhere to the intended semantic modifications, thereby demonstrating the effectiveness of the attack.

\item \textbf{Style transfer:} In style transfer tasks, we assess the impact of adversarial perturbations on the stylistic alignment between the denoised adversarial image \( D(\xnoi + \delta) \) and the target style image. We compute the {Gram matrix loss} \cite{gatys2015neural}, which measures the difference in style by comparing the correlations between feature maps (Gram matrices) of the two images extracted from a convolutional neural network's layers. The Gram matrix captures the texture and style information of an image. A lower Gram matrix loss indicates better alignment with the target style, suggesting that the adversarial perturbations have successfully enabled the denoised image to achieve the intended style transfer, thereby demonstrating their effectiveness.

\end{itemize}

\subsubsection{Overall evaluation}

Our comprehensive evaluation balances imperceptibility, clarity, and semantic impact. By ensuring that the adversarial perturbations are imperceptible (\( \text{PSNR}_{\text{con}} \), \( \text{SSIM}_{\text{con}} \), \( \text{LPIPS}_{\text{con}} \), and \( \text{Entropy}_{\text{con}} \)), we maintain the stealthiness of the attack. Simultaneously, high clarity metrics (PSNR, SSIM, LPIPS, and Entropy) between the denoised images and the original image confirm the effectiveness of the denoising process. Finally, semantic fidelity quantifies semantic modifications during the denoising process through downstream task evaluations, demonstrating the practical implications of our adversarial perturbations.

\subsection{Implementation details}
\label{sec:Implementation}
\subsubsection{Hyperparameter}
All experiments were conducted using NVIDIA V100 GPUs (32GB). Unlike traditional $\ell_p$-bounded attacks, MIGDA’s loss-driven approach does not require $\epsilon$ or $\alpha$. It typically converges within 30 iterations, taking approximately $0.8$ seconds on a $256 \times 256$ image. The Adam optimizer \cite{diederik2014adam} with an initial learning rate of \(10^{-4}\) was employed during the attack phase. We employed a learning rate scheduler with a step size of 30 and a decay factor (\(\gamma = 0.5\)) to dynamically adjust the learning rate during training.  
Through an extensive grid search process, the optimal set of hyperparameters was determined to be \(\alpha = 1\), \(\beta = 0.1\), \(\lambda_{\text{con}} = 3\), \(\lambda_{\text{rec}} = 1\), and \(\lambda_{\text{MI}} = 0.01\). Detailed experimental results and further discussion of these choices are presented in \cref{Hyp}.

\begin{table*}[ht]
   
    \centering
    \resizebox{\textwidth}{!}{ 
\begin{tabular}{c|c|cccccc|cccccc}
\hline
\multirow{2}{*}{Network}   & \multirow{2}{*}{Image} & \multicolumn{6}{c}{Tampered-IC13}             & \multicolumn{6}{c}{Synthetic Dataset}                               \\ \cline{3-14} 
                           &                        & PSNR$\uparrow$     & SSIM$\uparrow$ & LPIPS$\downarrow$ & Entropy$\downarrow$ & ROUGE-L$\uparrow$ &  & PSNR$\uparrow$     & SSIM$\uparrow$ & LPIPS$\downarrow$ & \multicolumn{1}{c}{Entropy$\downarrow$ }& ROUGE-L$\uparrow$ &  \\ \hline
\multirow{2}{*}{}          & $x$                    & $\infty$ & 1.00 & 0.00  & 6.60    & 0.57    &  & $\infty$ & 1.00 & 0.00  & \multicolumn{1}{c}{5.05}    & 0.15    &  \\
                           & $\xnoi$                & 23.98    & 0.47 & 0.48  & 7.32    & --      &  & 24.15    & 0.28 & 0.65  & \multicolumn{1}{c}{6.61}    & --      &  \\ \hline
\multirow{5}{*}{Restormer} & $D(\xnoi)$             & 27.89    & 0.87 & 0.29  & 6.65    & 0.57    &  & 28.22    & 0.89 & 0.42  & \multicolumn{1}{c}{5.28}    & 0.15    &  \\
                           & $x_{\text{adv}}$       & 25.58    & 0.56 & 0.46  & 7.05    & --      &  & 26.12    & 0.42 & 0.63  & \multicolumn{1}{c}{6.23}    & --      &  \\
                           & $D(x_{\text{adv}})$    & 13.92    & 0.38 & 0.58  & 6.52    & 0.32    &  & 17.65    & 0.62 & 0.63  & \multicolumn{1}{c}{5.44}    & 0.01    &  \\
                           & $x_{\text{MIGA}}$     & 23.06    & 0.45 & 0.48  & 7.32    & --      &  & 24.32    & 0.29 & 0.62  & \multicolumn{1}{c}{6.58}    & --      &  \\
                           & $D(x_{\text{MIGA}})$  & 24.95    & 0.85 & 0.17  & 6.75    & 0.77    &  & 34.09    & 0.94 & 0.17  & \multicolumn{1}{c}{5.05}    & 0.69    &  \\ \hline
\multirow{5}{*}{Xformer}   & $D(\xnoi)$             & 27.57    & 0.85 & 0.30  & 6.61    & 0.57    &  & 31.70    & 0.90 & 0.39  & \multicolumn{1}{c}{5.10}    & 0.15    &  \\
                           & $x_{\text{adv}}$       & 26.13    & 0.56 & 0.46  & 7.05    & --      &  & 26.54    & 0.42 & 0.62  & \multicolumn{1}{c}{6.23}    & --      &  \\
                           & $D(x_{\text{adv}})$    & 13.79    & 0.37 & 0.68  & 6.47    & 0.28    &  & 16.98    & 0.61 & 0.68  & \multicolumn{1}{c}{5.29}    & 0.00    &  \\
                           & $x_{\text{MIGA}}$     & 22.92    & 0.45 & 0.48  & 7.34    & --      &  & 24.07    & 0.28 & 0.65  & \multicolumn{1}{c}{6.61}    & --      &  \\
                           & $D(x_{\text{MIGA}})$  & 25.91    & 0.86 & 0.15  & 6.74    & 0.83    &  & 34.84    & 0.94 & 0.14  & \multicolumn{1}{c}{5.05}    & 0.75    &  \\ \hline
\multirow{5}{*}{PromptIR}  & $D(\xnoi)$             & 27.89    & 0.87 & 0.23  & 6.65    & 0.57    &  & 38.42    & 0.95 & 0.27  & \multicolumn{1}{c}{5.06}    & 0.15    &  \\
                           & $x_{\text{adv}}$       & 23.88    & 0.50 & 0.49  & 7.21    & --      &  & 24.11    & 0.35 & 0.66  & \multicolumn{1}{c}{6.49}    & --      &  \\
                           & $D(x_{\text{adv}})$    & 14.30    & 0.26 & 0.63  & 7.07    & 0.35    &  & 20.17    & 0.49 & 0.65  & \multicolumn{1}{c}{6.25}    & 0.01    &  \\
                           & $x_{\text{MIGA}}$     & 22.33    & 0.43 & 0.49  & 7.34    & --      &  & 23.56    & 0.27 & 0.65  & \multicolumn{1}{c}{6.63}    & --      &  \\
                           & $D(x_{\text{MIGA}})$  & 23.65    & 0.83 & 0.25  & 6.81    & 0.77    &  & 29.74    & 0.86 & 0.28  & \multicolumn{1}{c}{5.05}    & 0.54    &  \\ \hline
\multirow{5}{*}{AFM}  & $D(\xnoi)$             & 26.83            & 0.80             & 0.38              & 6.72                & 0.57                        &  & 32.53            & 0.87             & 0.21              & \multicolumn{1}{c}{5.41}                & 0.15                         &  \\
                           & $x_{\text{adv}}$       & 24.48            & 0.53             & 0.48              & 7.32                & --                          &  & 26.22            & 0.43             & 0.64              & \multicolumn{1}{c}{6.25}                & --                           &  \\
                           & $D(x_{\text{adv}})$    & 13.98            & 0.31             & 0.65              & 7.12                & 0.31                        &  & 17.32            & 0.64             & 0.69              & \multicolumn{1}{c}{5.52}                & 0.01                         &  \\
                           & $x_{\text{MIGA}}$      & 22.76            & 0.44             & 0.48              & 7.35                & --                          &  & 25.47            & 0.32             & 0.67              & \multicolumn{1}{c}{6.62}                & --                           &  \\
                           & $D(x_{\text{MIGA}})$   & 25.48            & 0.84             & 0.18              & 6.78                & 0.83                        &  & 35.72            & 0.93             & 0.15              & \multicolumn{1}{c}{5.27}                & 0.72                         &  \\ \hline 
\end{tabular}
    }
     \caption{Denoising attack for text alteration tasks.}
     \label{fig:u11}
\end{table*}

\begin{table*}[ht]
   
    \centering
    \resizebox{\textwidth}{!}{ 
        \begin{tabular}{c|c|cccccc|cccccc}
        \hline
        \multirow{2}{*}{Network}   & \multirow{2}{*}{Image} & \multicolumn{6}{c|}{MAGICBRUSH}                        & \multicolumn{6}{c}{Style Transfer}                      \\ \cline{3-14} 
                                   &                        & PSNR$\uparrow$     & SSIM$\uparrow$ & LPIPS$\downarrow$ & Entropy$\downarrow$ & CLIP similarity$\uparrow$ &  & PSNR$\uparrow$     & SSIM$\uparrow$ & LPIPS$\downarrow$ & Entropy$\downarrow$ & Gram matrix Loss$\downarrow$ &  \\ \hline
        \multirow{2}{*}{}          & $x$                    & $\infty$ & 1.00 & 0.00  & 7.36    & 0.23            &  & $\infty$ & 1.00 & 0.00  & 7.15    & 74.58            &  \\
                                   & $\xnoi$                & 34.56    & 0.94 & 0.07  & 7.40    & --              &  & 24.09    & 0.57 & 0.45  & 7.51    & --               &  \\ \hline
        \multirow{5}{*}{Restormer} & $D(\xnoi)$             & 37.43    & 0.97 & 0.05  & 7.37    & 0.23            &  & 26.57    & 0.83 & 0.32  & 7.09    & 21.98            &  \\
                                   & $x_{\text{adv}}$       & 25.26    & 0.62 & 0.45  & 7.46    & --              &  & 25.24    & 0.62 & 0.46  & 7.27    & --               &  \\
                                   & $D(x_{\text{adv}})$    & 13.22    & 0.28 & 0.63  & 7.16    & 0.22            &  & 12.57    & 0.16 & 0.68  & 6.89    & 327.53           &  \\
                                   & $x_{\text{MIGA}}$     & 31.64    & 0.93 & 0.08  & 7.40    & --              &  & 22.41    & 0.52 & 0.45  & 7.50    & --               &  \\
                                   & $D(x_{\text{MIGA}})$  & 26.07    & 0.93 & 0.05  & 7.38    & 0.24            &  & 21.45    & 0.56 & 0.25  & 7.31    & 13.95            &  \\ \hline
        \multirow{5}{*}{Xformer}   & $D(\xnoi)$             & 36.60    & 0.97 & 0.05  & 7.36    & 0.23            &  & 25.56    & 0.79 & 0.34  & 7.03    & 23.75            &  \\
                                   & $x_{\text{adv}}$       & 25.91    & 0.63 & 0.44  & 7.47    & --              &  & 25.91    & 0.63 & 0.45  & 7.28    & --               &  \\
                                   & $D(x_{\text{adv}})$    & 13.31    & 0.28 & 0.72  & 6.97    & 0.22            &  & 13.16    & 0.16 & 0.73  & 6.74    & 418.85           &  \\
                                   & $x_{\text{MIGA}}$     & 31.14    & 0.92 & 0.08  & 7.40    & --              &  & 21.76    & 0.52 & 0.45  & 7.52    & --               &  \\
                                   & $D(x_{\text{MIGA}})$  & 26.75    & 0.94 & 0.05  & 7.37    & 0.24            &  & 21.00    & 0.55 & 0.21  & 7.22    & 9.10             &  \\ \hline
        \multirow{5}{*}{PromptIR}  & $D(\xnoi)$             & 42.27    & 0.99 & 0.03  & 7.36    & 0.23            &  & 30.13    & 0.91 & 0.21  & 7.13    & 110.77           &  \\
                                   & $x_{\text{adv}}$       & 23.48    & 0.55 & 0.48  & 7.57    & --              &  & 23.56    & 0.55 & 0.48  & 7.40    & --               &  \\
                                   & $D(x_{\text{adv}})$    & 12.98    & 0.23 & 0.62  & 7.37    & 0.22            &  & 12.85    & 0.12 & 0.66  & 7.33    & 1467.96          &  \\
                                   & $x_{\text{MIGA}}$     & 30.00    & 0.91 & 0.09  & 7.40    & --              &  & 21.40    & 0.49 & 0.47  & 7.48    & --               &  \\
                                   & $D(x_{\text{MIGA}})$  & 25.65    & 0.93 & 0.05  & 7.38    & 0.24            &  & 20.78    & 0.59 & 0.28  & 7.42    & 52.88            &  \\ \hline
\multirow{5}{*}{AFM}       & $D(\xnoi)$             & 36.21            & 0.95             & 0.06              & 7.35                & 0.23                        &  & 25.53            & 0.71             & 0.33              & 7.09                & 23.32                        &  \\
                           & $x_{\text{adv}}$       & 25.87            & 0.62             & 0.45              & 7.47                & --                          &  & 24.33            & 0.63             & 0.47              & 7.28                & --                           &  \\
                           & $D(x_{\text{adv}})$    & 13.33            & 0.29             & 0.70              & 7.08                & 0.22                        &  & 13.22            & 0.17             & 0.69              & 6.73                & 455.92                       &  \\
                           & $x_{\text{MIGA}}$      & 30.69            & 0.92             & 0.07              & 7.40                & --                          &  & 21.29            & 0.49             & 0.45              & 7.54                & --                           &  \\
                           & $D(x_{\text{MIGA}})$   & 27.10            & 0.93             & 0.08              & 7.37                & 0.24                        &  & 20.15            & 0.51             & 0.25              & 7.26                & 15.34                        &  \\ \hline
        \end{tabular}}
        \caption{Denoising attack for content replacement and style transfer tasks.}
         \label{fig:u22}
\end{table*}

\begin{table*}[h]

    \centering
    \resizebox{\textwidth}{!}{ 
\begin{tabular}{c|c|c|c|c|c|c}
\hline
\multicolumn{1}{l}{Model}                       & Metric                                  & \multicolumn{1}{l|}{ImageNet-10} & \multicolumn{1}{l|}{Tampered-IC13} & \multicolumn{1}{l|}{Synthetic Dataset} & \multicolumn{1}{l|}{MAGICBRUSH} & \multicolumn{1}{l}{Style Transfer} \\ \hline
\multicolumn{1}{c|}{\multirow{3}{*}{Restormer}} & $\text{PSNR}_{\text{con}}$$\uparrow$    & 30.38/38.53                      & 31.26/28.11                        & 31.75/35.66                            & 30.96/33.82                     & 30.80/25.87                        \\
\multicolumn{1}{c|}{}                           & $\text{SSIM}_{\text{con}}$$\uparrow$    & 0.98/0.99                        & 0.95/0.92                          & 0.93/0.94                              & 0.94/0.97                       & 0.94/0.87                          \\
\multicolumn{1}{c|}{}                           & $\text{LPIPS}_{\text{con}}$$\downarrow$ & 0.04/0.01                        & 0.07/0.02                          & 0.11/0.01                              & 0.08/0.01                       & 0.08/0.04                          \\ \hline
\multicolumn{1}{c|}{\multirow{3}{*}{Xformer}}   & $\text{PSNR}_{\text{con}}$$\uparrow$    & 30.51/41.56                      & 32.50/29.00                        & 33.21/35.66                            & 32.31/33.34                     & 32.16/25.10                        \\
\multicolumn{1}{c|}{}                           & $\text{SSIM}_{\text{con}}$$\uparrow$    & 0.98/0.99                        & 0.94/0.94                          & 0.93/0.97                              & 0.88/0.97                       & 0.94/0.88                          \\
\multicolumn{1}{c|}{}                           & $\text{LPIPS}_{\text{con}}$$\downarrow$ & 0.04/0.01                        & 0.08/0.02                          & 0.12/0.01                              & 0.08/0.01                       & 0.08/0.04                          \\ \hline
\multicolumn{1}{c|}{\multirow{3}{*}{PromptIR}}  & $\text{PSNR}_{\text{con}}$$\uparrow$    & 28.97/34.39                      & 28.46/26.65                        & 28.01/30.30                            & 27.58/31.40                     & 27.74/23.51                        \\
\multicolumn{1}{c|}{}                           & $\text{SSIM}_{\text{con}}$$\uparrow$    & 0.97/0.99                        & 0.92/0.90                          & 0.89/0.91                              & 0.89/0.96                       & 0.89/0.82                          \\
\multicolumn{1}{c|}{}                           & $\text{LPIPS}_{\text{con}}$$\downarrow$ & 0.07/0.01                        & 0.15/0.04                          & 0.27/0.03                              & 0.18/0.02                       & 0.18/0.08                          \\ \hline
\multicolumn{1}{c|}{\multirow{3}{*}{AFM}}       & $\text{PSNR}_{\text{con}}$$\uparrow$    & 31.48/39.82                      & 30.02/28.16                        & 32.24/34.47                            & 27.58/28.26                     & 28.64/24.49                        \\
\multicolumn{1}{c|}{}                           & $\text{SSIM}_{\text{con}}$$\uparrow$    & 0.98/0.99                        & 0.93/0.92                          & 0.92/0.96                              & 0.87/0.94                       & 0.91/0.85                          \\
\multicolumn{1}{c|}{}                           & $\text{LPIPS}_{\text{con}}$$\downarrow$ & 0.02/0.01                        & 0.09/0.02                          & 0.11/0.01                              & 0.19/0.02                       & 0.12/0.03                          \\ \hline
\end{tabular}
    }
     \caption{Perturbation constraint results. The values before the slash represent the constraint magnitude of traditional I-FGSM attack methods compared to the original images, while the values after the slash represent the constraint magnitude of MIGA.}
     \label{tab:constraint}
\end{table*}

\begin{figure*}[ht]
    \centering
    \includegraphics[width=1.0\textwidth]{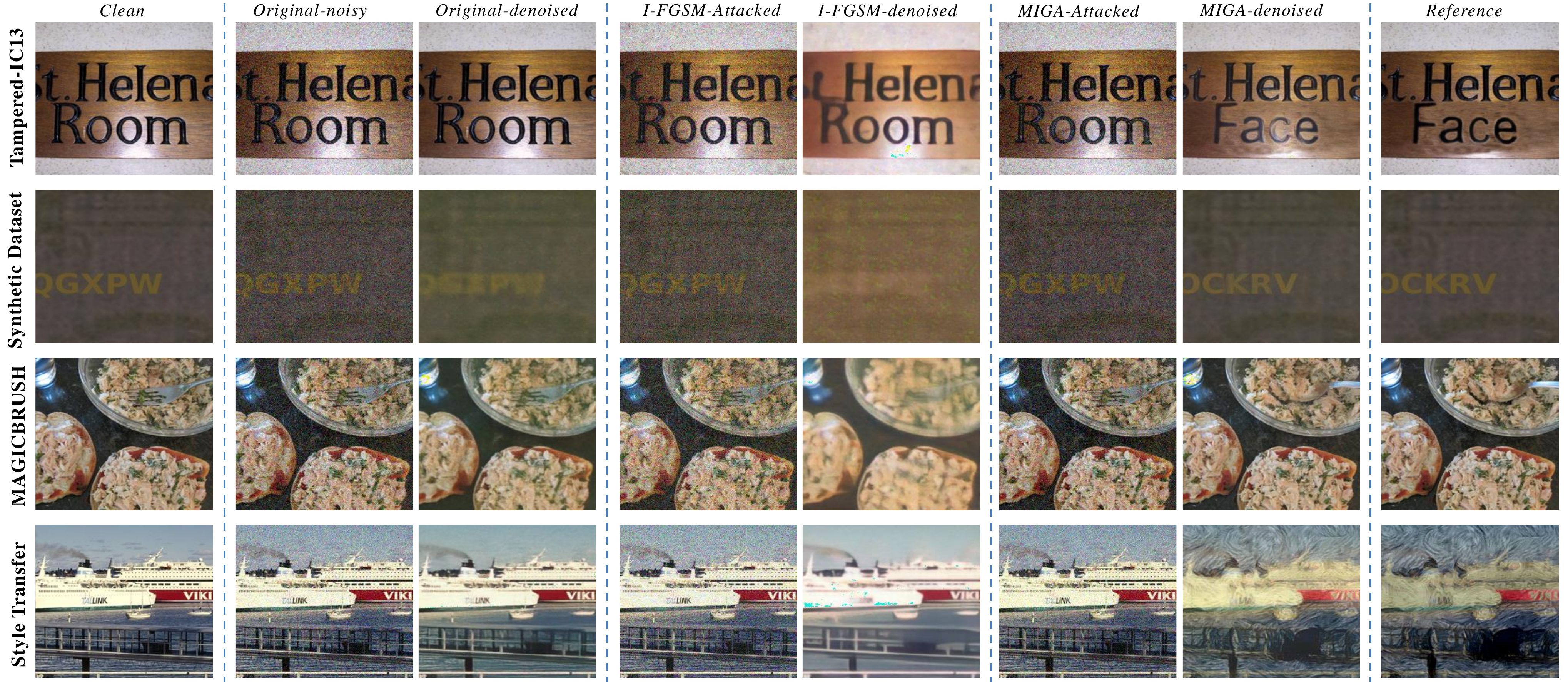} 
    \caption{Attack results on unknown tasks. Traditional attacks like I-FGSM degrade image quality, while MIGA modifies key semantic information in the denoised images generated after the attack.}
\vspace{-0.4cm}

    \label{fig:dd2}
\end{figure*}

\subsubsection{Training of MINE network}
\label{sec:mine_training}
For unknown downstream tasks, we used the MINE network to estimate the task-related mutual information. 
Specifically, we trained a neural network estimator \(T\) based on the MINE framework \cite{belghazi2018mine}, where the positive sample pair consists of the original image \(x\) and the denoised image \(D(\xnoi)\), and the negative sample pair consists of the original image \(x\) and a semantically altered image \(x_{\text{alt}}\). The goal is to let  the network \(T\)  distinguish between positive and negative sample pairs to estimate the mutual information value for a given image pair.

The training process uses a contrastive loss function:
\begin{multline}
L_{\text{contrastive}} \\
= -\log \left( \frac{\exp(T(x, D(\xnoi)))}{\exp(T(x, D(\xnoi))) + \exp(T(x, x_{\text{alt}}))} \right).
\end{multline}
This loss function encourages the network to assign higher similarity scores to positive sample pairs, while giving lower scores to negative sample pairs. After training, \(T(x, D(\xnoi))\) represents the similarity between the original image \(x\) and the denoised image, reflecting the semantic relevance of the image.

\begin{figure*}[h]
    \centering
    \includegraphics[width=0.75\textwidth]{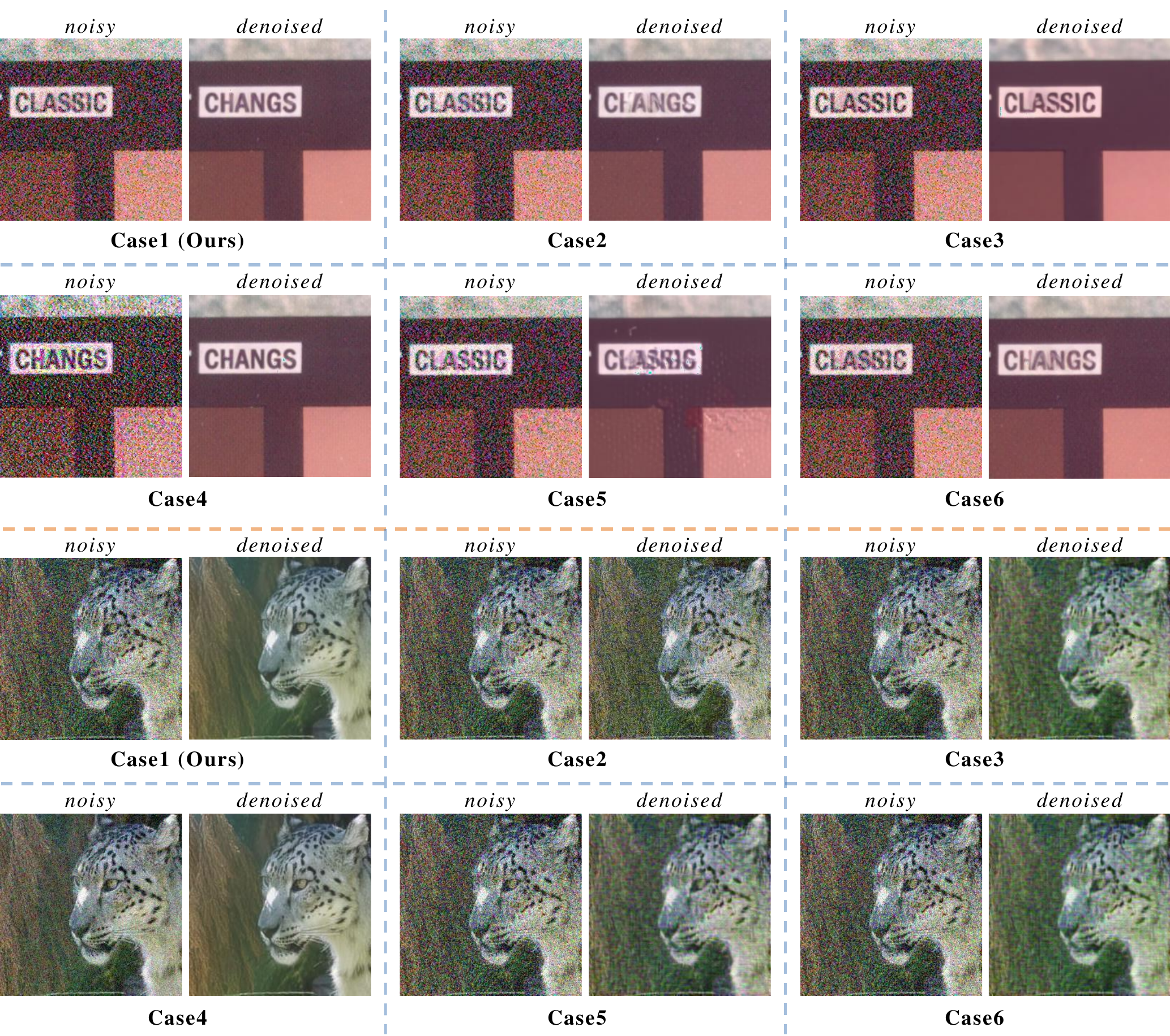} 
    \vspace{-0.2cm}
    \caption{Attack results of MIGA on the Tampered-IC13 (top) and ImageNet-10 (bottom) datasets under different loss combinations.Specifically, \textit{case1} through \textit{case6} sequentially correspond to the experimental settings listed from top to bottom in \cref{known weight}.}
\vspace{-0.4cm}
    \label{fig:ab}
\end{figure*}
\begin{table}[ht]

    \centering
    \setlength{\tabcolsep}{2pt} 
    \renewcommand{\arraystretch}{0.9} 
    \resizebox{\columnwidth}{!}{ 
    \begin{tabular}{cc|cc|c|cccc}
        \toprule
        \multicolumn{2}{c|}{$L {\text{com}}$} & \multicolumn{2}{c|}{$L {\text{rec}}$} & \multirow{2}{*}{$L_{\text{MI}}$} & \multicolumn{4}{c}{Metrics} \\
        \cmidrule(lr){1-2} \cmidrule(lr){3-4} \cmidrule(lr){6-9}
        MSE & $L {\text{perc}}$ & MSE & $L {\text{perc}}$ &  & LPIPS(con) & LPIPS & Entropy & Accuracy \\
        \midrule
        \checkmark & \checkmark & \checkmark & \checkmark & \checkmark & 0.01 & 0.26 & 7.06 & 51.54 \\ 
        \checkmark & \checkmark & \checkmark & \checkmark & $\times$ & 0.01 & 0.26 & 7.36 & 86.04 \\
        \checkmark & \checkmark & $\times$ & $\times$ & \checkmark & 0.01 & 0.37 & 7.41 & 52.08 \\
        $\times$ & $\times$ & \checkmark & \checkmark & \checkmark & 0.01 & 0.36 & 7.41 & 51.46 \\
        \checkmark & $\times$ & \checkmark & $\times$ & \checkmark & 0.01 & 0.36 & 7.40 & 52.19 \\
        $\times$ & \checkmark & $\times$ & \checkmark & \checkmark & 0.01 & 0.36 & 7.41 & 51.58 \\
        \bottomrule
    \end{tabular}}
    \caption{Importance of different losses for known task.}
    \label{known weight}
\end{table}

\begin{figure*}[ht]
    \centering
    \includegraphics[width=0.75\textwidth]{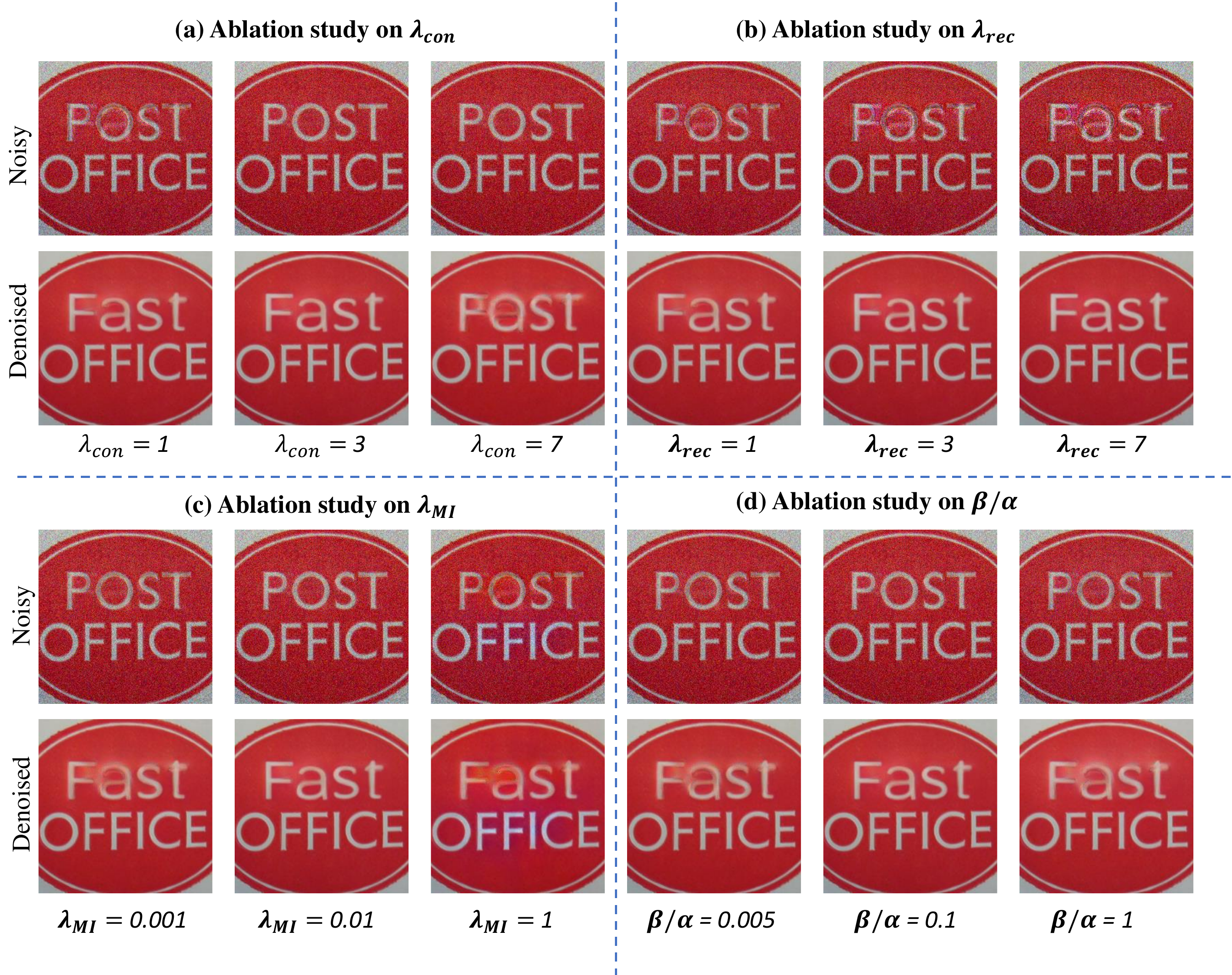} 
    \vspace{-0.4cm}
    \caption{Hyperparameter selection results.}
\vspace{-0.4cm}
    \label{fig:dd4}
\end{figure*}

\section{Experimental results} 
\label{more results}

\subsection{Denoising attack for unknown downstream tasks}

In this section, we further evaluate the performance of MIGA in scenarios where the downstream tasks are unknown to the attacker.  \cref{fig:u11} and \cref{fig:u22} present a comprehensive comparison between our proposed method MIGA and the traditional adversarial attack method I-FGSM across multiple state-of-the-art denoising models, including Restormer, Xformer, PromptIR, and AFM. The evaluation is conducted on several datasets, namely Tampered-IC13, Synthetic Dataset, MAGICBRUSH, and Style Transfer tasks, to ensure the generality of our findings.
{\cref{tab:constraint}} quantifies the degree of perturbations introduced to the original images by both methods. Specifically, it reports the perceptual similarity metrics such as $\text{PSNR}_{\text{con}}$, $\text{SSIM}_{\text{con}}$, and $\text{LPIPS}_{\text{con}}$ between the adversarial examples and the original noisy  images. Additionally, \cref{fig:dd2} provides visual illustrations of the denoised images after the attacks, highlighting the perceptual quality and the effectiveness of the perturbations in altering the semantics of the denoised image.

Consistent with the conclusions drawn from 
Sec. 5.3, our results demonstrate that MIGA effectively introduces imperceptible perturbations that remain undetectable to the human eye while significantly impacting the performance of downstream tasks after denoising. For instance, when using the Restormer network on the Tampered-IC13 dataset, the perceptual metric \(\text{LPIPS}_{\text{con}}\) increases only marginally by 0.02, indicating minimal perceptual difference from the original image. However, the ROUGE-L score, which measures the quality of text recognition or generation tasks, significantly improves to 0.77, reflecting the success of the attack in altering the downstream task outcome.
In contrast, images attacked by I-FGSM show a substantial decrease in PSNR and SSIM values, leading to noticeable degradation in image quality. This degradation not only makes the perturbations perceptible but also could raise suspicion in real-world applications where image integrity is crucial. Our method, therefore, presents a more stealthy and effective approach for attacking denoising models in scenarios where the attacker lacks knowledge of the specific downstream tasks.

\subsection{Importance of loss components}
\label{Importance of loss}

To further understand the contribution of each component in our loss function design, we add ablation studies under the scenario where the downstream tasks are known. The experimental results are detailed in  \cref{known weight}, which lists various combinations of the loss components and their corresponding performance metrics.  \cref{fig:ab} visually compares the outcomes under different ablation settings, providing qualitative insights into how each loss term affects the final results. Specifically, \textit{case1} through \textit{case6} sequentially correspond to the experimental settings listed from top to bottom in \cref{known weight}.
Our comprehensive analysis reveals that the inclusion of all loss components in our method yields the best performance. Specifically, incorporating both the content loss (\(L_{\text{con}}\)) and the reconstruction loss (\(L_{\text{rec}}\)) alongside the mutual information loss (\(L_{\text{MI}}\)) ensures that the adversarial perturbations are minimal, the denoised images are clear, and the semantic content is effectively altered. 

Notably, when certain loss components are omitted, the performance degrades. For example, \textit{case4}, which excludes the content loss, results in clear denoised images but lacks constraints on the perturbations, leading to significant alterations in the original images. This is evident in the Tampered-IC13 dataset results, where semantic content is visibly changed even before denoising. Such changes could be easily detected, compromising the stealthiness of the attack. Therefore, our loss function design is crucial for balancing the imperceptibility of the perturbations with the effectiveness of the attack.

\subsection{Hyperparameter selection}
\label{Hyp}
The selection of hyperparameters in our algorithm plays a pivotal role in balancing the trade-offs between perturbation imperceptibility, denoised image clarity, and semantic alteration of the downstream tasks. The hyperparameters \(\alpha\), \(\beta\), \(\lambda_{\text{con}}\), \(\lambda_{\text{rec}}\), and \(\lambda_{\text{MI}}\) correspond to the weights of different loss components and constraints in our optimization objective.
As discussed in  \cref{Importance of loss}, \(\lambda_{\text{con}}\) regulates the magnitude of the perturbations added to the original images, ensuring that they remain imperceptible. \(\lambda_{\text{rec}}\) influences the reconstruction fidelity of the denoised images, affecting their visual clarity. \(\lambda_{\text{MI}}\) controls the degree to which the mutual information between the denoised images and the original images is minimized, thereby altering the semantic content relevant to the downstream tasks. The ratio of \(\alpha\) to \(\beta\) determines the balance between pixel-level fidelity and perceptual similarity in the constraint loss.

To empirically determine the optimal hyperparameters, we performed a grid search over a range of values, systematically evaluating the impact of each parameter on the attack's effectiveness and imperceptibility.  \cref{fig:dd4} presents a subset of the results from our hyperparameter tuning experiments. 
Our analysis indicates that the optimal settings are \(\alpha = 1\), \(\beta = 0.1\), \(\lambda_{\text{con}} = 3\), \(\lambda_{\text{rec}} = 1\), and \(\lambda_{\text{MI}} = 0.01\). These values achieve a harmonious balance, ensuring that the perturbations are imperceptible, the denoised images are visually clear, and the downstream tasks are effectively disrupted. These settings were consistently effective across different datasets and denoising models, demonstrating the robustness of our approach.

\subsection{Results of other advanced aderveral attack method-CosPGD}

\begin{table}[t]

    \centering
   
    \setlength{\tabcolsep}{2pt} 
    \renewcommand{\arraystretch}{0.9} 
    \resizebox{\columnwidth}{!}{ 
    \begin{tabular}{c|c|cccc|ccccc|ccccc}
        \toprule
        Network & Image & PSNR$\uparrow$ & SSIM$\uparrow$ & LPIPS$\downarrow$ & Entropy$\downarrow$ & ROUGE-L$\uparrow$ \\
        \midrule
         \multirow{11}{*}{{Xformer}} 
        & $x$    & $\infty$ & 1.00 & 0.00 & 6.60 & 0.57 \\
        & $\xnoi$  & 23.98 & 0.47 & 0.48 & 7.32 & --  \\
        & $D(\xnoi)$   & 27.57 & 0.85 & 0.30 & 6.61 & 0.57 \\
        & $x_{\text{adv}}$       & 26.13    & 0.56 & 0.46  & 7.05    & --     \\
                           & $D(x_{\text{adv}})$    & 13.79    & 0.37 & 0.68  & 6.47    & 0.28     \\

         & $x_{\text{CosPGD(original)}}$ & 27.22  & 0.60 & 0.42  & 7.03 &-- \\      
        & $D(x_{\text{CosPGD(original)}})$ & 14.94  & 0.41 & 0.52 & 6.48 &0.01 \\
       
         & $x_{\text{CosPGD(adapted)}}$ & 24.06  & 0.48 & 0.46 & 7.29 &-- \\      
        & $D(x_{\text{CosPGD(adapted)}})$ & 22.80  &  0.78 & 0.28 &  6.52 &0.12 \\
                             & $x_{\text{MIGA}}$     & 22.92    & 0.45 & 0.48  & 7.34    & --      \\
                           & $D(x_{\text{MIGA}})$  & 25.91    & 0.86 & 0.15  & 6.74    & 0.83     \\ \hline
        \bottomrule
    \end{tabular}
    }
    \vspace{-0.3cm}
\caption{Comparison of different adversarial attacks on the Tampered-IC13 dataset. }
\label{tab:cos}
    \vspace{-0.3cm}
\end{table}

\begin{table}[t]

    \centering
    \resizebox{\columnwidth}{!}{ 
    \begin{tabular}{c|cccc}
    \toprule
    \multicolumn{1}{c|}{Metric} & \multicolumn{1}{l}{I-FGSM} & \multicolumn{1}{l}{CosPGD(original} & \multicolumn{1}{l}{CosPGD(adapted} & \multicolumn{1}{l}{MIGA} \\ 
    \midrule

    \text{$\text{LPIPS}_{\text{con}}$$\downarrow$} & 0.04  & 0.01  & 0.02  & 0.01  \\ 
    \bottomrule
    \end{tabular}
    }
    \vspace{-0.3cm}
    \caption{Comparison of perturbation constraints on the Tampered-IC13 task.  
}

    \vspace{-0.5cm}
    
    \label{tab:constraint3}
\end{table}
We extend our evaluation by comparing MIGA with CosPGD~\cite{agnihotri2023cospgd}, a pixel-level adversarial attack. While CosPGD primarily degrades image clarity, it fails to manipulate semantics effectively when applied to denoising models.  
To explore its potential for semantic modification, we adapt CosPGD by introducing semantically modified target images.  
As shown in~\cref{tab:cos}, the original CosPGD lowers PSNR without altering semantics, whereas the adapted version maintains image clarity but still struggles to induce meaningful semantic shifts.  
Unlike CosPGD, which remains an untargeted attack focusing on PSNR degradation, MIGA explicitly manipulates semantic consistency while preserving visual fidelity, achieving a more deceptive and effective attack.  
For instance, MIGA significantly outperforms CosPGD in semantic manipulation, achieving a ROUGE-L score of 0.83, while even the adapted CosPGD fails to induce substantial semantic changes.  
Furthermore, \cref{tab:constraint3} ensures a fair comparison by verifying that all attack methods introduce imperceptible perturbations to the original image.

\subsection{Additional results on robustness against defenses}
\begin{figure}[t]
    \centering
    \includegraphics[width=0.5\textwidth]{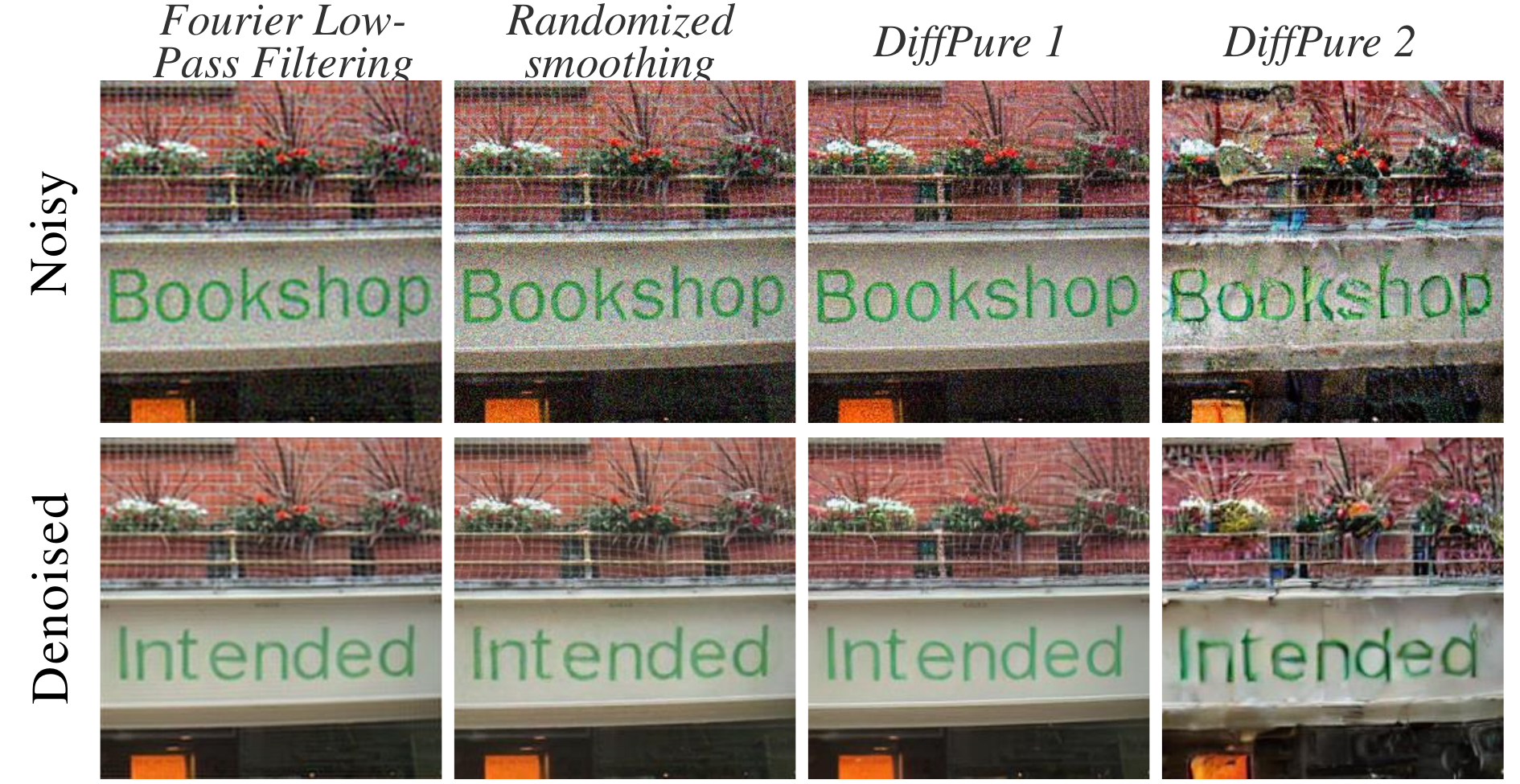} 
    
    \caption{Impact of different defense strategies on MIGA. Fourier Low-Pass Filtering, Randomized Smoothing, and two levels of DiffPure (DiffPure 1: strength=0.05, DiffPure 2: strength=0.3) are applied to the denoising process. Despite stronger defenses, MIGA-induced semantic shifts (e.g., from ``Bookshop'' to ``Intended'') persist, highlighting the robustness of our attack.}
    \vspace{-0.3cm}
    \label{fig:roub_append}
\end{figure}
We further evaluate the resilience of MIGA adversarial examples against various defense strategies, including Fourier Low-Pass Filtering~\cite{deng2024exploring}, Randomized Smoothing~\cite{chiang2020certified}, and DiffPure~\cite{nie2022diffusion} with different strengths. 
As shown in ~\cref{fig:roub_append}, while these defenses can mitigate certain adversarial effects, MIGA remains effective in altering semantic information post-denoising. Notably, {DiffPure 2}, which applies stronger modifications to the original image (with a strength parameter of 0.3), still fails to fully remove the adversarial effect. The denoised output continues to exhibit a semantic shift, such as from \textit{``Bookshop''} to \textit{``Intended''}, demonstrating the robustness of MIGA.
These findings align with our previous results in main paper's Tab. 8 and Fig. 5, where MIGA-induced semantic changes persist despite defense applications. This confirms that MIGA remains a persistent threat even under strong defense mechanisms.

\subsection{Transferability on other denoising models}

An important aspect of adversarial attacks is their transferability—the ability of adversarial examples generated for one model to be effective against other models. To evaluate this, we tested the adversarial examples generated using one denoising model on other models, examining both the degradation in downstream task performance and the preservation of image quality.
As summarized in Table \cref{transfer}, our adversarial examples exhibit significant transferability across different denoising architectures. For instance, adversarial examples crafted using the Xformer model and tested on the Restormer model resulted in a classification accuracy of 35.92\%, which is comparable to the accuracy of 35.23\% when tested on the original Xformer model. This indicates that the adversarial perturbations are not denoising model-specific and can generalize to other architectures.

Moreover, the image quality metrics such as PSNR and SSIM after denoising remain high, suggesting that the transfer does not substantially degrade the visual quality of the images. This is crucial for practical scenarios where the adversarial examples need to remain inconspicuous to human observers.

\begin{table}[t]

    \centering
    \resizebox{\columnwidth}{!}{ 
    \begin{tabular}{c|cccc|c}
        \hline
        Transfer  &PSNR$\uparrow$ &SSIM$\uparrow$ & LPIPS$\downarrow$ & Entropy$\downarrow$ & Accuracy$\downarrow$ \\
        \midrule
        Origin & 17.65 & 0.30 & 0.62 & 7.71 & 83.00 \\
        No-Attack & 22.36 & 0.70 & 0.38 & 6.81 & 89.85 \\
        
        Xformer-Xformer & 25.64 & 0.67 & 0.22 & 7.13 & 35.23 \\
        Xformer-Restormer & 24.95 & 0.77 & 0.32 & 7.36 & 35.92 \\
        PromptIR-PromptIR & 24.20 & 0.59 & 0.37 & 7.60 & 21.31 \\
        PromptIR-Restormer & 24.94 & 0.66 & 0.38 & 7.36 & 31.04 \\
        \hline
    \end{tabular}}
    \vspace{-0.3cm}
    \caption{Transfer evaluation: Xformer-Restormer represents adversarial samples generated against Xformer and tested on Restormer.}
    \label{transfer}
    \vspace{-0.5cm}
\end{table}

\subsection{More experimental results visualization}

\begin{figure*}[ht]
    \centering
    \includegraphics[width=1.0\textwidth]{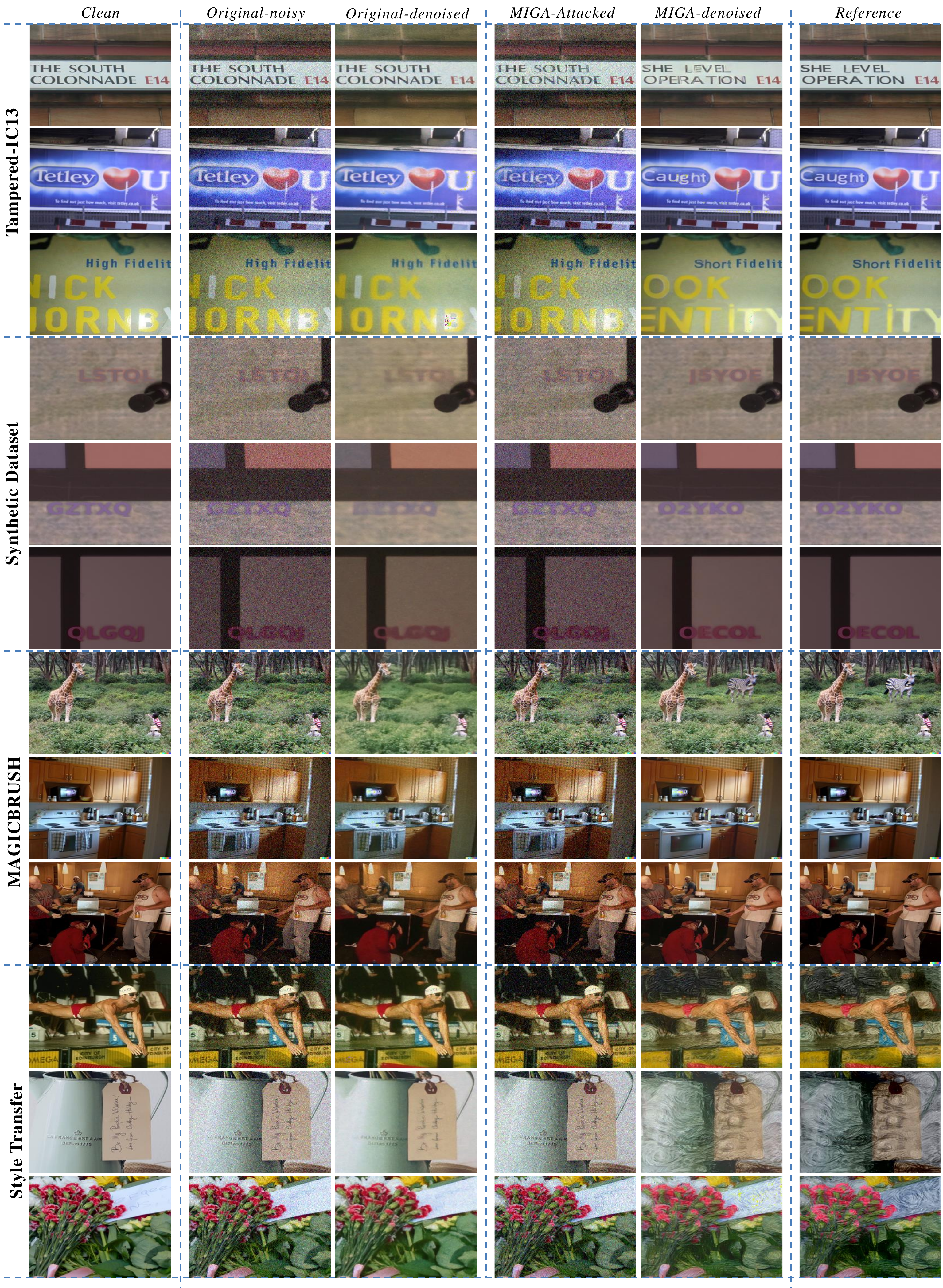} 
    \caption{Visualization of MIGA results on Tampered-IC13, Synthetic Dataset, MAGICBRUSH, and Style Transfer.}

    \label{fig:dd3}
\end{figure*}

To provide a more comprehensive understanding of the effectiveness of our proposed method, we present more visualizations of the attack results on various datasets. Figure \cref{fig:dd3} showcases examples from the Tampered-IC13, Synthetic Dataset, MAGICBRUSH, and Style Transfer tasks. These visualizations illustrate how MIGA effectively alters the semantic content of the denoised images in a way that impacts the downstream tasks while keeping the perturbations imperceptible.

In the Tampered-IC13 and Synthetic Dataset examples, the text content in the images is subtly altered post-denoising, leading to significant errors in text recognition tasks. For the MAGICBRUSH and Style Transfer tasks, the stylistic attributes or content of the images are modified, affecting the outcomes of tasks like image synthesis or style transfer. These results demonstrate the versatility and robustness of our method on denoising models with different types of downstream tasks.
Furthermore, they underscore the critical need for developing more resilient denoising techniques and robust defenses to counteract such adversarial attacks.

\end{document}